\documentclass[10pt,letterpaper,conference]{ieeeconf}
\usepackage[bookmarks=false,hidelinks=true]{hyperref}
\usepackage{times}
\usepackage{graphicx,dblfloatfix}
\usepackage{mathrsfs,amsmath}
\usepackage{amssymb}
\usepackage{mathtools}
 
\usepackage{float}
\usepackage{listings}
\usepackage{color}
\usepackage{xcolor}
\usepackage{caption}
\usepackage{subfigure}
\usepackage{amsmath}
\usepackage{algorithm}
\usepackage{algpseudocode}
\usepackage{epigraph}
\usepackage{booktabs}
\usepackage{setspace}
\usepackage{amsfonts,amssymb}
\usepackage{multirow}

\IEEEoverridecommandlockouts
\title{\LARGE \bf Learning monocular visual odometry with dense 3D mapping \\
from dense 3D flow}

\author{Cheng Zhao$^{1}$, Li Sun$^{2}$, Pulak Purkait$^{3}$, Tom Duckett$^{2}$ and Rustam Stolkin$^{1}$
\thanks{$^{1}$ Extreme Robotics Lab, University of Birmingham, Birmingham, UK, B15 2TT.
        {\tt\small IRobotCheng@gmail.com}. $^{2}$ Lincoln Centre for Autonomous Systems (L-CAS), University of Lincoln, UK, LN6 7TS. $^{3}$ Cambridge Research Lab, Toshiba Research Europe, Cambridge, UK, CB4 0GZ.}
}
       
\begin{document}              
\maketitle
\thispagestyle{empty}
\pagestyle{empty}
\begin{abstract}
This paper introduces a fully deep learning approach to monocular SLAM, which can perform simultaneous localization using a neural network for learning visual odometry (L-VO) and dense 3D mapping. Dense 2D flow and a depth image are generated from monocular images by sub-networks, which are then used by a 3D flow associated layer in the L-VO network to generate dense 3D flow. Given this 3D flow, the dual-stream L-VO network can then predict the 6DOF relative pose and furthermore reconstruct the vehicle trajectory. In order to learn the correlation between motion directions, the Bivariate Gaussian modeling is employed in the loss function. The L-VO network achieves an overall performance of $2.68\%$ for average translational error and $0.0143 ^\circ/m$ for average rotational error on the KITTI odometry benchmark.  Moreover, the learned depth is leveraged to generate a dense 3D map. As a result, an entire visual SLAM system, that is, learning monocular odometry combined with dense 3D mapping, is achieved. 
\end{abstract}
\section{Introduction}\label{sec:1}
Simultaneous localization and mapping~(SLAM) is an essential technique for mobile robot applications. In the past few decades, a substantial amount of research has been devoted to visual SLAM systems that enable robots to localize robustly and accurately in different environments. One of the most challenging branches of visual SLAM is monocular SLAM, which often suffers critically from absolute scale drift. Usually, some prior knowledge such as the height of the camera is necessary to alleviate scale drift. Moreover, these methods require hand-coded engineering efforts and excellent parameter tuning skills.

In recent years, deep learning techniques for visual odometry and SLAM have attracted considerable attention in the SLAM community. These methods not only provide good performance in challenging environments but also rectify the incorrect scale estimation of monocular SLAM. Supervised learning approaches formulate visual odometry (VO) as a regression problem. They explore the ability of CNN~\cite{costante2017ls} or RNN~\cite{wang2017end}\cite{clark2017vinet} to learn ego-motion estimation using the change of RGB value features~\cite{haarnoja2016backprop}, deep flow~\cite{muller2017flowdometry} and non-deep flow~\cite{costante2016exploring} features. These methods are calibration-free but require a lot of expensive ground truth data for training. 

On the other hand, some networks for predicting VO take advantage of geometric constraints, e.g.\ similarity constraints, epipolar constraints, etc., by integrating them into the loss function and training the network in an unsupervised manner. Although the trajectory ground truth is not required for these methods, consecutive frames~\cite{zhou2017unsupervised}\cite{ummenhofer2017demon}\cite{vijayanarasimhan2017sfm} or stereo image pairs~\cite{li2017undeepvo} along with the above geometric constraints are enough to provide sufficient supervision to train the network. However, these methods usually require the intrinsic parameters of the camera. 

\begin{figure}[t]
	\centering
	\includegraphics[width= 0.4\textwidth]{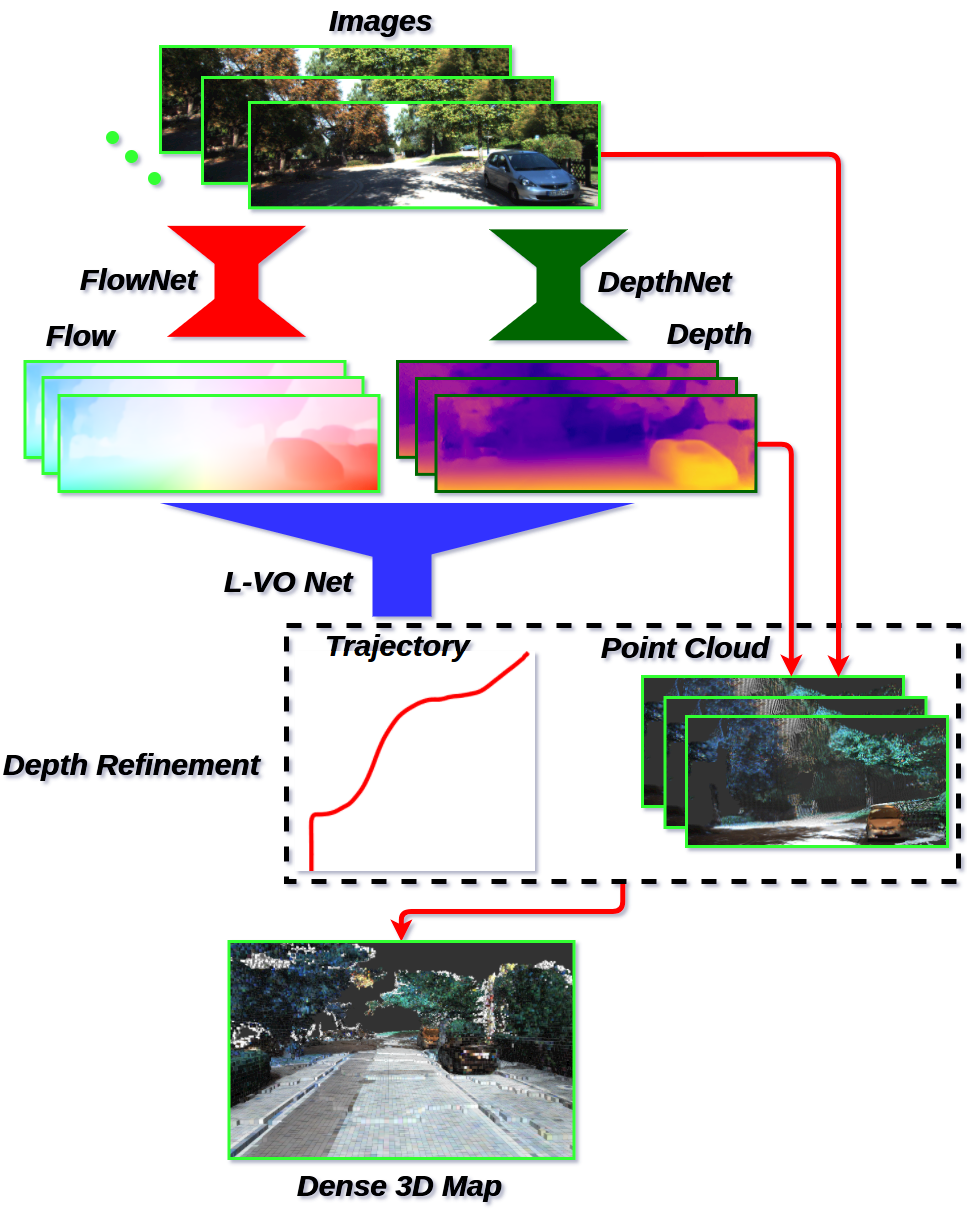}
	\caption{The pipeline of the proposed learning monocular SLAM system. More detail can be found in Sec.~\ref{sec:3.1}. 
}
	\label{fig:DeepVO_Mapping}
\end{figure}

The main limitation of the above methods is that they all suffer from high dataset bias and require domain similarity between the training and testing sequences. Moreover, most of the deep learning geometry research only focus on visual odometry for localization without mapping. CNN-SLAM~\cite{tateno2017cnn} is the forerunner to integrate learning of depth prediction with monocular SLAM to generate an accurate dense 3D map. But the odometry in CNN-SLAM is still based on the conventional method. Therefore, it is still not a pure deep learning SLAM method. In addition, some researches~\cite{zhao2017fully}\cite{zhao2017dense}\cite{sun2017weakly}\cite{sun2018recurrent} integrate deep semantic information into a conventional SLAM system.     

In this paper, 
a learning system for monocular SLAM is developed, 
which can simultaneously perform 
localization and dense 3D mapping through an end-to-end neural network. A learning visual odometry (L-VO) network with a 3D association layer is proposed for ego-motion estimation, which achieves an overall performance of $2.68\%$ for average translational error and $0.0143 ^\circ/m$ for average rotational error on the KITTI\footnote{\url{http://www.cvlibs.net/datasets/kitti}} odometry benchmark. 
The main contributions can be 
briefly summarized as follows: 
i)~A new baseline L-VO method with a 3D association layer is proposed for ego-motion estimation, 
ii)~a Bivariate Gaussian loss function is used to learn the correlation between motion directions,  
iii)~L-VO is extended to a learning monocular SLAM system. 
An overview of the proposed architecture is shown in Fig.~\ref{fig:DeepVO_Mapping}.  

\section{Related work}\label{sec:2}

\subsection{Learning based visual odometry (Pre-deep learning era)}\label{sec:2.1}
 In the recent past, some learning-based visual odometry estimation methods~\cite{roberts2008memory}\cite{roberts2009learning}\cite{guizilini2011visual}\cite{guizilini2012semi}\cite{ciarfuglia2014evaluation} were explored,  before deep learning began to dominate many computer vision and robotics tasks. These learning-based methods mainly explored different pre-deep learning methods such as SVM, Gaussian Processes, etc.\ using sparse optical flow features for camera localisation and motion estimation.  

\subsection{Supervised deep learning for visual odometry}\label{sec:2.2}
One of the pioneering works on deep learning for visual odometry estimation was proposed by Costante \emph{et al.}~\cite{costante2016exploring}. They employed convolutional neural networks (CNNs) for ego-motion estimation from dense optical flow obtained by a non-deep method~\cite{brox2004high}. 
Then, Muller \emph{et al.}~\cite{muller2017flowdometry} proposed Flowdometry, which combines FlowNet~\cite{dosovitskiy2015flownet} and CNNs to obtain an end-to-end odometry system. Gabriele \emph{et al.}~\cite{costante2017ls} proposed Latent Space Visual Odometry (LS-VO) to find a non-linear representation of the optical flow manifold.  

Tuomas \emph{et al.}~\cite{haarnoja2016backprop} explored LSTM for visual odometry. They utilized CNNs on the temporal change of RGB values (temporal derivatives) between two adjacent images. They utilized LSTM as a baseline in their work and proposed a back-propagation method for a Kalman Filter to learn the discriminative deterministic state estimators. Another seminal work on learning visual odometry was proposed by Wang \emph{et al.}~\cite{wang2017end}\cite{wang2017deepvo}. They utilized FlowNet features with LSTM for an end-to-end visual odometry system. Clark \emph{et. al}~\cite{clark2017vinet} used the same network but fused the features of the monocular RGB camera with additional IMU readings for improved performance. Mehmet \emph{et al.}~\cite{turan2018deep} adopted a similar architecture~-- CNNs with LSTM~-- to develop a visual odometry system for endoscopic capsule robots. 

\subsection{Unsupervised deep learning for visual odometry}\label{sec:2.3}
Most of the unsupervised visual odometry estimation methods predict the depth and ego-motion simultaneously. These methods do not require the trajectory ground truth but need camera parameters and often some additional information such as stereo images for training. Benjamin \emph{et al.}~\cite{ummenhofer2017demon} proposed the DeMoN architecture, which estimates not only depth and motion but also the surface normals and optical flow from a pair of images. They employed an unsupervised training loss function based on the relative spatial differences. Zhou \emph{et al.}~\cite{zhou2017unsupervised} also used a training loss function which minimizes image warping error of an image sequence for unsupervised depth prediction and ego-motion estimation. SfM-Net~\cite{vijayanarasimhan2017sfm} predicts depth, segmentation, camera and rigid object motions, and transforms these to obtain frame-to-frame dense optical flow. Li \emph{et al.}~\cite{li2017undeepvo} combined the loss functions from \cite{zhou2017unsupervised} and \cite{godard2017unsupervised} to obtain an unsupervised visual odometry method that can recover the absolute scale.  

\subsection{Learning visual SLAM}\label{sec:2.4}
Most of the deep learning geometry research only focuses on VO for localization without mapping. The only forerunner of deep learning SLAM, CNN-SLAM~\cite{tateno2017cnn}, integrates CNN-style depth prediction with monocular SLAM to recover the absolute scale, and meanwhile generates a dense 3D map. 
However the odometry in CNN-SLAM is still based on the conventional method. 
As the estimated odometry of CNN-SLAM is not based on learning methods, it is not a complete end-to-end approach for learning SLAM.    

\subsection{Discussion}\label{sec:2.5}
Conventional monocular visual odometry suffers from scale drift. Pioneering researchers~\cite{muller2017flowdometry}\cite{wang2017deepvo}\cite{clark2017vinet}\cite{costante2017ls} show that this problem can be mitigated via learning from 2D flow features. Inspired by RGBD-SLAM, the relative transform can be estimated directly from solving the PnP problem when the depth is given. In this paper, we model the visual odometry problem as a probabilistic regression problem. Multi-modal features, i.e.\ 3D flow (derived from the 2D flow and depth flow), are used to enhance the observation of the learning visual odometry. We further explore the correlation of motion directions and learn the translation with a multi-variate Gaussian rather than isotropic Gaussian~\cite{wang2017end}. Moreover, the learned depth is leveraged to generate a dense 3D map. As a result, an entire visual SLAM system, that is, learning monocular odometry combined with dense 3D mapping, is achieved.  
\section{Methodology}\label{sec:3}
  
\subsection{Overview}\label{sec:3.1}
The pipeline of the proposed learning monocular SLAM is shown in Fig.~\ref{fig:DeepVO_Mapping}. The proposed L-VO network is an end-to-end neural network for simultaneous monocular visual odometry and dense 3D mapping. To be more specific, L-VO Net takes a pair of consecutive images as input and predicts Ego-motion. The dense 2D flow and depth are obtained with FlowNet2~\cite{ilg2017flownet} and DepthNet~\cite{godard2017unsupervised} respectively. The estimated dense 2D flow and depth are further associated to obtain the 3D flow. Next, the 3D flow is fed into two separate regressors to predict the 6DOF relative pose (including scale) transform between each pair of images. As a consequence, the 6DOF camera trajectory can be obtained by accumulating relative poses. The point cloud is simultaneously generated and mapped incrementally from the given RGB image and the estimated depth. Furthermore, a 3D refinement is employed to remove the outliers and incorrect predictions. Finally, a dense 3D map is generated. 


\begin{figure*}[thpb]
	\centering
	\includegraphics[width= 1.0\textwidth]{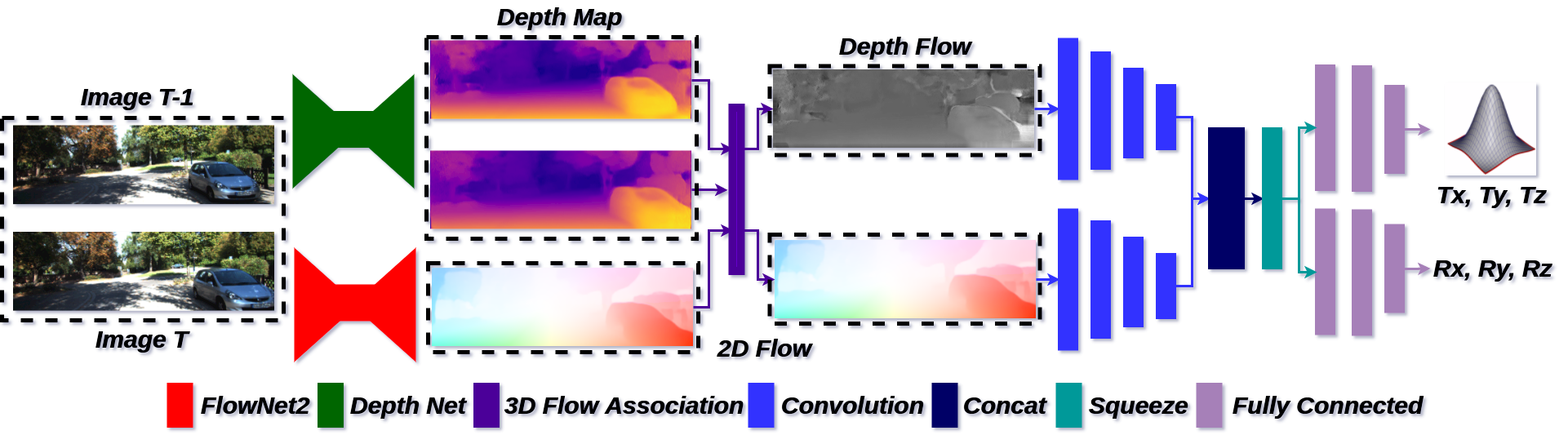}
	\caption{The architecture of the proposed learning visual odometry (L-VO) network.}
	\label{fig:DeepVO_Net}
\end{figure*}

\subsection{2D optical flow and depth prediction}\label{sec:3.2}
For dense 2D optical flow prediction, the state-of-the-art approach FlowNet2~\cite{ilg2017flownet} is employed. FlowNet2 is a stacked architecture composed of a series of FlowNet-S~\cite{dosovitskiy2015flownet}, FlowNet-C~\cite{dosovitskiy2015flownet} and FlowNet-SD~\cite{ilg2017flownet}. 
It can deliver robust 2D dense optical flow, which is of significant importance for learning odometry. We fine-tune this network using the training KITTI data (as described in \ref{sec:4.3}) and then transplant the network for our task.

For depth prediction, any of the state-of-the-art methods~\cite{ummenhofer2017demon}, \cite{zhou2017unsupervised} and \cite{godard2017unsupervised} can be adapted to our approach. In this paper, \cite{godard2017unsupervised} is employed because of its good performance in outdoor scenes. \cite{godard2017unsupervised} is an encoder-decoder architecture with appearance matching loss, disparity smoothness loss and left-right disparity consistency loss, which can be trained in unsupervised fashion. The training objective enables the network to perform the depth estimation from a monocular image. The network is also fine-tuned using the training KITTI data (as described in \ref{sec:4.3}).    

\subsection{3D flow association layer}\label{sec:3.3}
We propose a 3D flow association layer which generates dense 3D flow from 2D flow and the corresponding depth maps. Assuming $F_{XY}^{k:k+1} \in \mathbb{R}^{h\times w \times2}$ is the predicted dense 2D flow (on X-Y image plane) between frame $k$ and $k+1$, and $D^{k}\in \mathbb{R}^{h\times w}$ is the predicted depth map of frame $k$, the 3D flow association layer can be defined as:



\begin{equation}
F_Z^{k:k+1} (x,y) = D^{k+1} \big( (x,y)+F_{XY}^{k:k+1}(x,y) \big) - D^{k}(x,y)
\end{equation}


\begin{equation}
F_{3D}^{k:k+1} = \mathcal{C}(F_{XY}^{k:k+1}, F_Z^{k:k+1})
\end{equation}

\noindent where $F_{3D}^{k:k+1} (x,y) \in \mathbb{R}^{3}$ refers to the 3D flow at pixel coordinate $(x,y)$ and $\mathcal{C}$ is the concatenation operation. If the depth value in frame $k+1$ cannot be associated with the corresponding depth value in frame $k$, the missing flow pixels between two adjacent frames can be interpolated through bilinear filtering. It is worth noting that the inverse depth (i.e.\ disparity) is more sensitive to the motion of surroundings and objects close to the camera. Hence, the inverse depth is used instead of the depth value in our approach. We still use the term ``depth'' in order to make the following description more readable. 


\subsection{Learning odometry}\label{sec:3.4}
As shown in Fig.~\ref{fig:DeepVO_Net}, our learning odometry network is a dual stream architecture network, composed of two branches of convolution stacks followed by a squeeze layer~\cite{iandola2016squeezenet} and two fully connected regressors. The convolution layers are composed of $3\times3$ filters and are of stride $2$. The numbers of channels in the two branches are $64$, $128$, $256$ and $512$. In order to keep the spatial geometry information, the pooling layer is abandoned in these two CNN stacks. In the end, the feature maps of the two branches are concatenated together and squeezed using a $1\times1$ filter: 
\begin{equation}
F_{3D} = \mathcal{S}(F_{XY}, F_{Z}) 
\end{equation}
\noindent where $\mathcal{S}$ is the squeeze operation, $F_{XY} \in \mathbb{R}^{h\times w\times n}$ and $F_{Z} \in \mathbb{R}^{h\times w\times n}$ are the feature maps of 2D flow and depth flow respectively,  $F_{3D} \in \mathbb{R}^{h\times w\times n/4}$ is the squeezed feature, and $n$ is the number of feature channels. The squeeze layer embeds the 3D feature map into a lower dimensional space, thereby reducing the input dimension of the regressors. A triple-layer fully-connected network is used for regression. We set the hidden layers of the regressors to size $128$ with $relu$ activation function. The output of the translation regressor is $6$ for bivariate Gaussian loss and that of the rotation regressor is $3$, which is trained through a $\ell_2$ loss. The details of the loss function are described as follows. 

\subsection{Bivariate Gaussian loss function}\label{sec:3.5}
For most of the outdoor on-road driving data, e.g. KITTI dataset, there is a strong correlation between the translations along different axes in the horizontal plane. In contrast with the previous loss functions used in learning odometry, we aim to let our network learn the correlation along the forward and left/right translation directions. In this paper, this correlation is modeled as a multivariate Gaussian distribution.

The same camera configuration~(axes definitions) as in the KITTI dataset is used, i.e. $x:$ right (horizontal), $y:$ down (vertical), $z:$ forward (horizontal), then the translation variation along $y$ coordinate is small compared to the other axes. Therefore, we only need to find the correlations between translation $x$ and translation $z$. In our approach, the Bivariate Gaussian Probabilistic-Density-Function ($PDF$)~\cite{graves2013generating} is employed as the likelihood function for $x$ (left/right) and $z$ (forward) translation prediction. For the translation in $y$ direction and orientations, $\ell_2$ loss is used for optimization. Similar to~\cite{elgammal2004inferring}, the Euler angles rather than quaternion are used to represent the orientation, as the quaternion representation opens up the possibility of over-fitting in the rotation regression. We further include a $\ell_2$ regularization term for all weights to mitigate over-fitting. Our loss function is defined as: 
\begin{equation}
\begin{split}
loss = \sum_i^N -\log\left(PDF((x_{gt}^i, z_{gt}^i),~\mathcal{N}^i(\mu, \Sigma)\right) ~~~~~~~\\ 
+ \lambda_{1} \sum_i^N \Vert y_p^i-y_{gt}^i\Vert_2 + \lambda_{2} \sum_i^N  \Vert r_p^i-r_{gt}^i\Vert_2 + \lambda_{3} \Vert W \Vert_2 &
\end{split}
\end{equation}
\noindent where $N$ is the number of training pair images, $(x_{gt}^i, y_{gt}^i, z_{gt}^i)$ is the ground truth camera translation, and $(x_p^i, y_p^i, z_p^i)$ is the predicted translation of the $i^{th}$ image/camera. $r_p^i:=(e_p^z, e_p^y, e_p^x)^i$ and $r_{gt}^i:=(e_{gt}^z, e_{gt}^y, e_{gt}^x)^i$ are the predicted and ground-truth Euler angles, respectively. $W$ are the trainable weights of the neural network. $\lambda_{1}$, $\lambda_{2}$ and $\lambda_{3}$ are the scale factors to balance the weights of translation and orientations. 
The Gaussian Density Function $PDF$ is defined as:
\begin{equation}
\begin{split}
PDF\left((x_{gt}^i, z_{gt}^i),~\mathcal{N}^i(\mu, \Sigma)\right) =  ~~~~~~~~~~~~~~~~\\
\frac{exp(-\frac{1}{2}((x_{gt}^i, z_{gt}^i)-\mu) \Sigma^{-1} ((x_{gt}^i, z_{gt}^i)-\mu)^{T})} {((2 \pi)^2|\Sigma|)^{-1/2}}
\end{split}
\end{equation}

\noindent where the bivariate Gaussian distribution $\mathcal{N}$ is:  

\begin{equation}
\small
\mu = ( \mu_x, \mu_z )^i
, ~ \Sigma = \Big( 
\begin{tabular}{cc} 
  $\sigma_x^2$ & $\rho \sigma_x \sigma_z$ \\
  $\rho \sigma_x \sigma_z$ & $\sigma_z^2$,
\end{tabular}
\Big)^i. 
\end{equation}

\noindent where $\mu_x$ and $\mu_z$ are two mean variables in the left/right and forward direction, $\sigma_x, \sigma_z$ are the corresponding standard deviations and $\rho$ is the correlation coefficient of the translation between left/right and forward direction in the horizontal plane. Our neural network is expected to learn $(\mu_x, \mu_z, \sigma_x, \sigma_z, \rho, y_p)$, and $(e_p^z, e_p^y, e_p^x)$, corresponding to the 6-dimensional and 3-dimensional outputs of two regression neural networks. 

Once the network is trained, i.e.\ the translation $(\mu_x, \mu_z, \sigma_x, \sigma_z, \rho, y_p)$ and rotation $(e_p^z, e_p^y, e_p^x)$ can be estimated from the network, the predicted translation in the horizontal plane is obtained through sampling within the bi-variant Gaussian distribution using:

\begin{equation}
x, z = \frac{1}{N_{s}} \sum_k^{N_{s}} (x_s, z_s)^k \sim \mathcal{N}_{p}(\mu, \Sigma),
\end{equation}  

\noindent where $\mathcal{N}_p$ is obtained from $(\mu_x, \mu_z, \sigma_x, \sigma_z, \rho)$, $(x_s, z_s)^k$ is the $k$th sample, and $N_s$ is the number of samples.

\subsection{Octree depth fusion for mapping}\label{sec:3.6} 
We also proposed a dense 3D mapping method using the learned odometry and depth. Given the RGB image and the corresponding predicted depth image, the 3D point cloud $(X, Y, Z)$ can be obtained through:
 
\begin{equation}\label{eq:backprojection}
d_{u, v}~ 
\begin{bmatrix}
    u\\
    v\\
    1
\end{bmatrix} =
\begin{bmatrix}
    f_{x}~s~c_{x}\\
    0~f_{y}~c_{y}\\
    0~~0~~1\\
\end{bmatrix}
~
\begin{bmatrix}
    X\\
    Y\\
    Z
\end{bmatrix}
\end{equation}
\noindent where $f_{x}, f_{y}$ are the focal lengths, $(c_{x}, c_{y})$ is the principal point offset and $s$ is the axis skew. $(u, v)$ is the pixel position in the image plane.

Unfortunately, the depth prediction usually suffers from blur around the depth borders. The predicted depth is not accurate enough to be utilized directly for 3D mapping. In our approach, the OctoMap representation~\cite{hornung2013octomap} is used to refine and maintain the 3D map. In order to build a robust, accurate dense 3D map, depth fusion using measurements from multiple views is employed. In OctoMap, each leaf node $n$ stores the occupancy probability $P(n|o_{1:t})$. Given the 3D point measurements $o_{1:t}$, the probability $P(n|o_{1:t})$ can be updated as:
\vspace{-1mm}
\begin{equation}
\small
P(n|o_{1:t}) = \left[1 + \dfrac{1-P(n|o_{t})}{P(n|o_{t})} \dfrac{1-P(n|o_{1:t-1})}{P(n|o_{1:t-1})} \dfrac{P(n)}{1-P(n)} \right]^{-1}
\end{equation}

\noindent here, $P(n|o_{t})$ can be obtained by a beam tracing sensor model. If the probability $P(n|o_{1:t})$ of the leaf node is beyond a threshold, this node will be marked as occupied in the dense 3D map. This probabilistic occupancy fusion can fuse the depth estimations from multiple views, and remove points arising from inaccurate depth predictions.   

\newcommand{\sizeOfimage}{0.3}
\begin{figure*}[thpb]
\centering
\subfigure[Sequence 03. \label{fig:sequence03}]{\includegraphics[width= \sizeOfimage\textwidth]{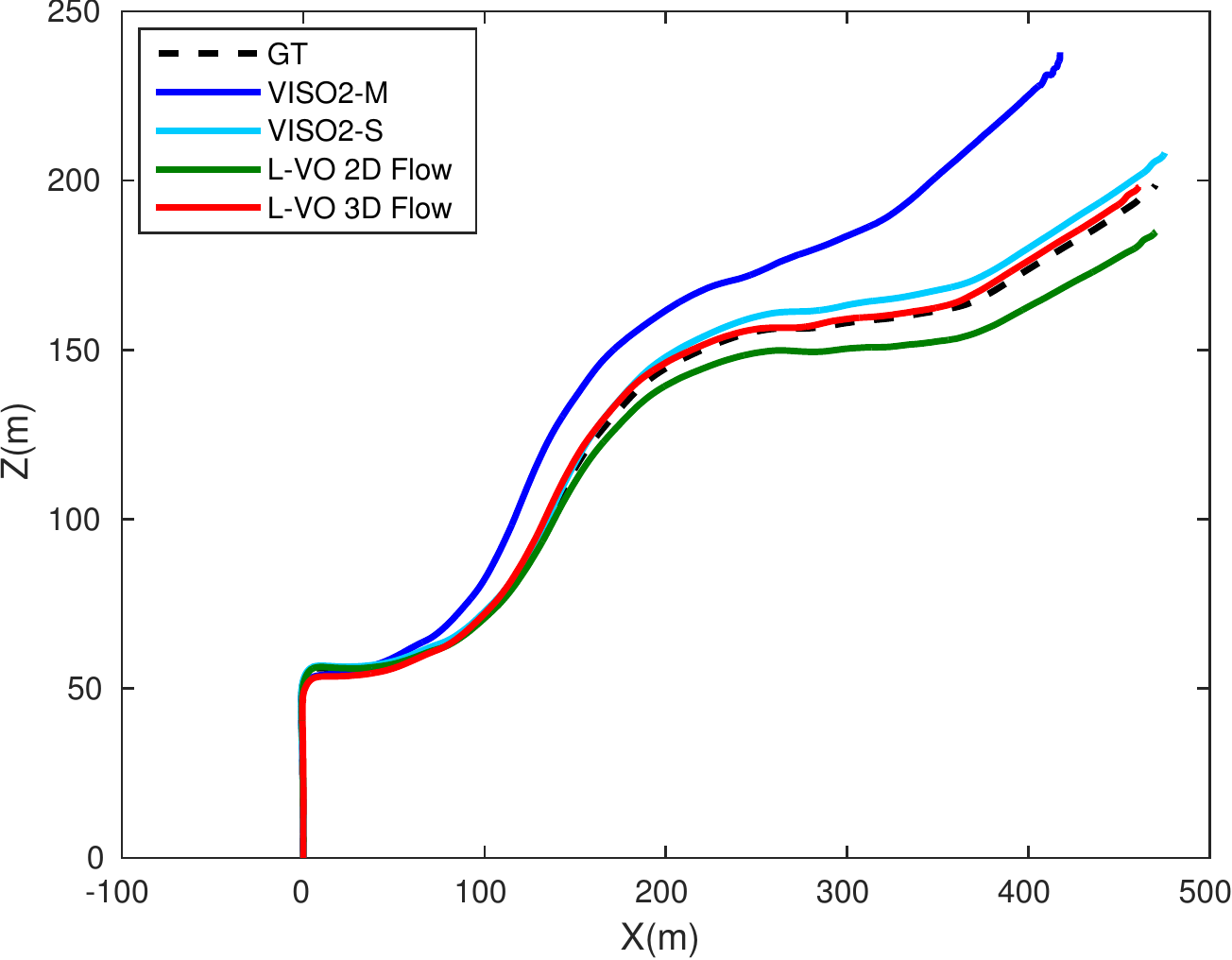}}
\subfigure[Sequence 04. \label{fig:sequence04}]{\includegraphics[width= \sizeOfimage\textwidth]{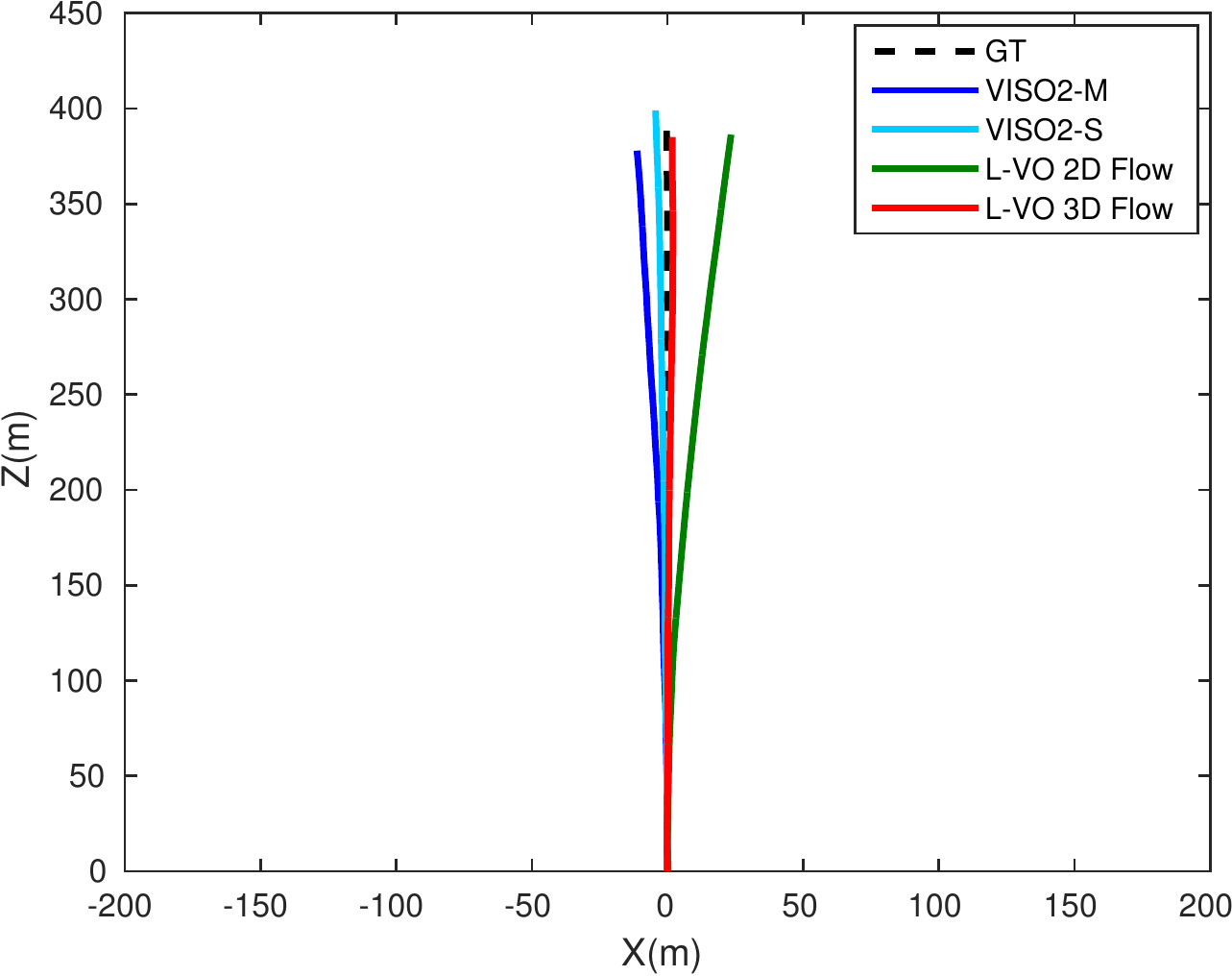}}
\subfigure[Sequence 05. \label{fig:sequence05}]{\includegraphics[width= \sizeOfimage\textwidth]{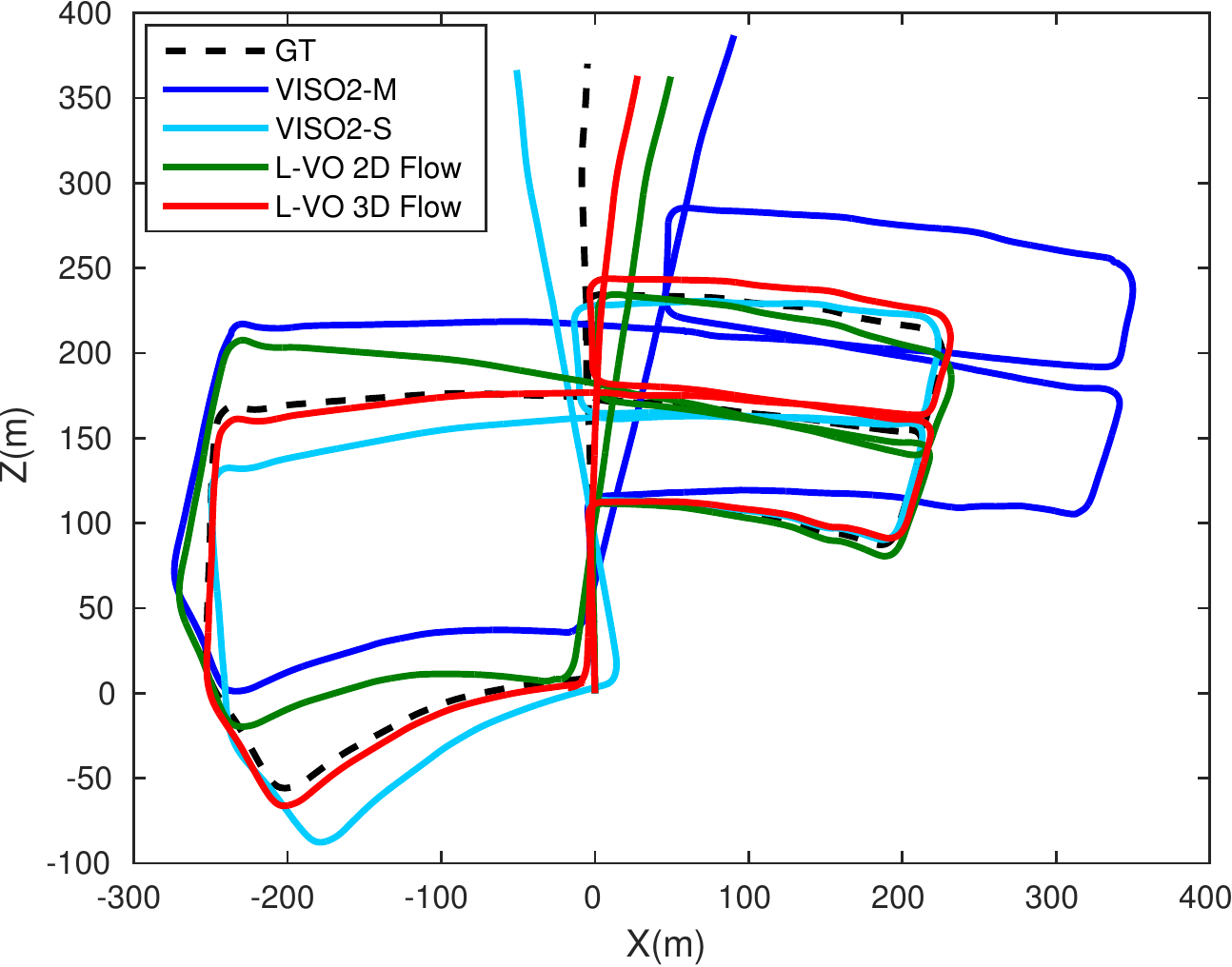}}
\subfigure[Sequence 06. \label{fig:sequence06}]{\includegraphics[width= \sizeOfimage\textwidth]{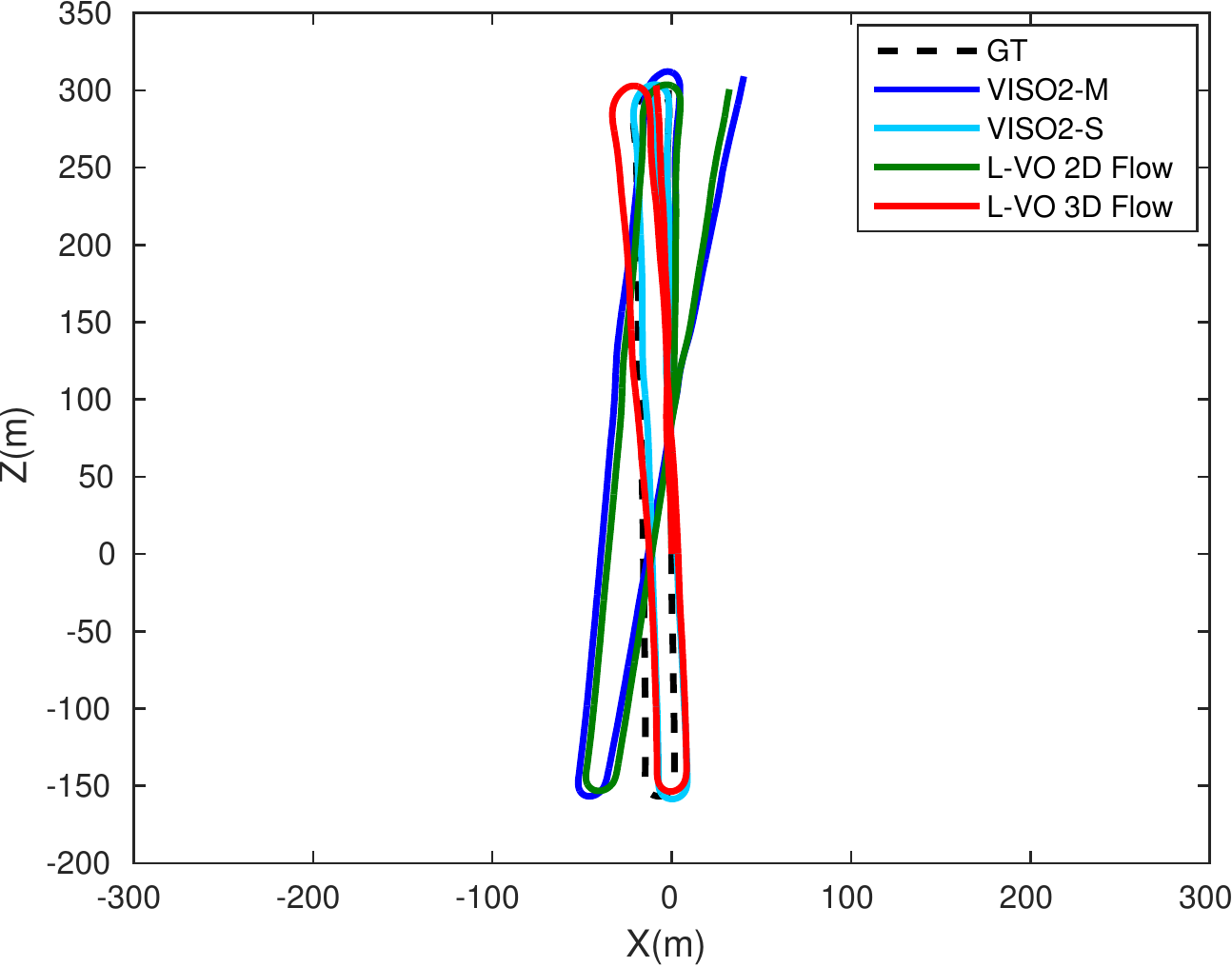}}
\subfigure[Sequence 07. \label{fig:sequence07}]{\includegraphics[width= \sizeOfimage\textwidth]{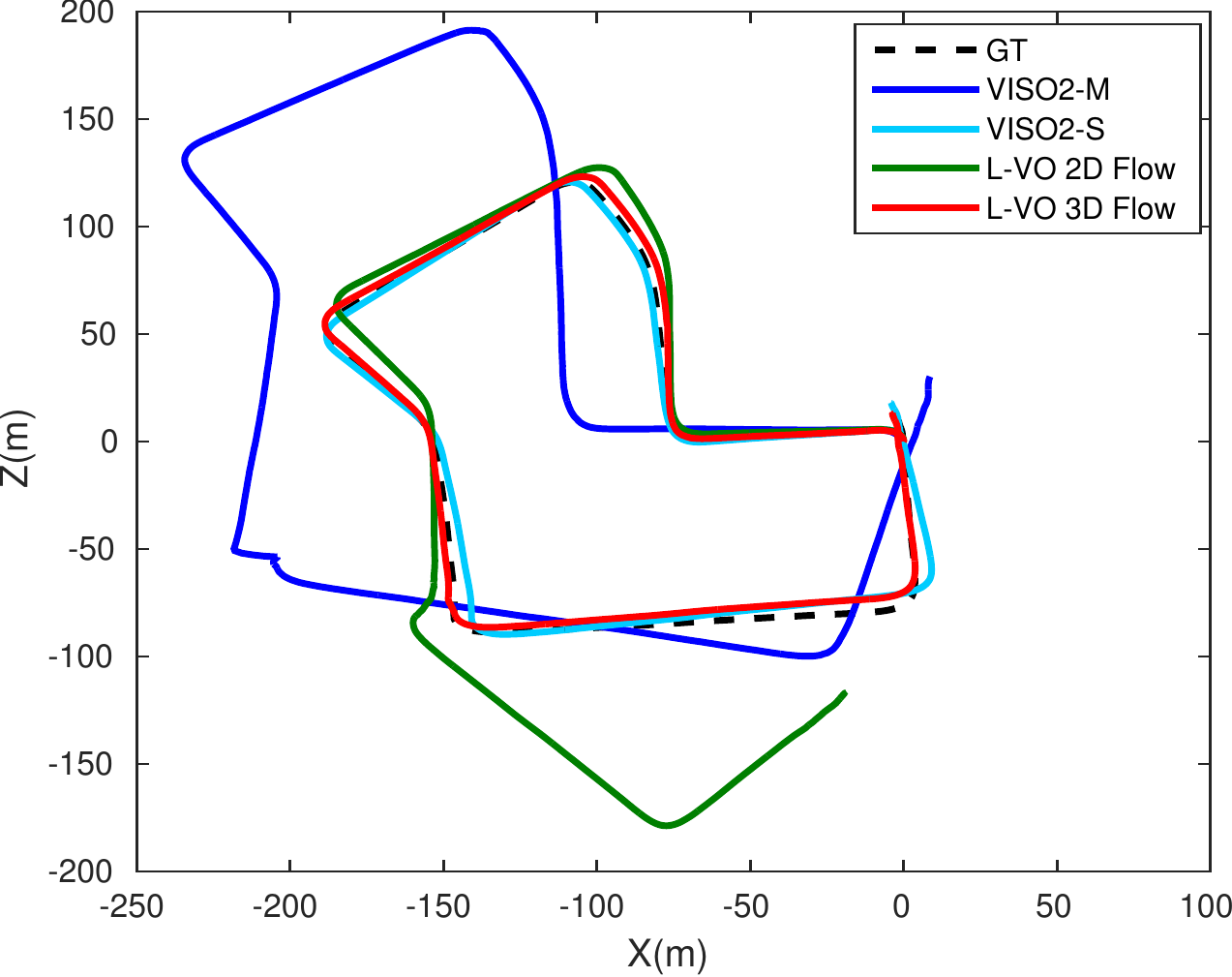}}
\subfigure[Sequence 10. \label{fig:sequence10}]{\includegraphics[width= \sizeOfimage\textwidth]{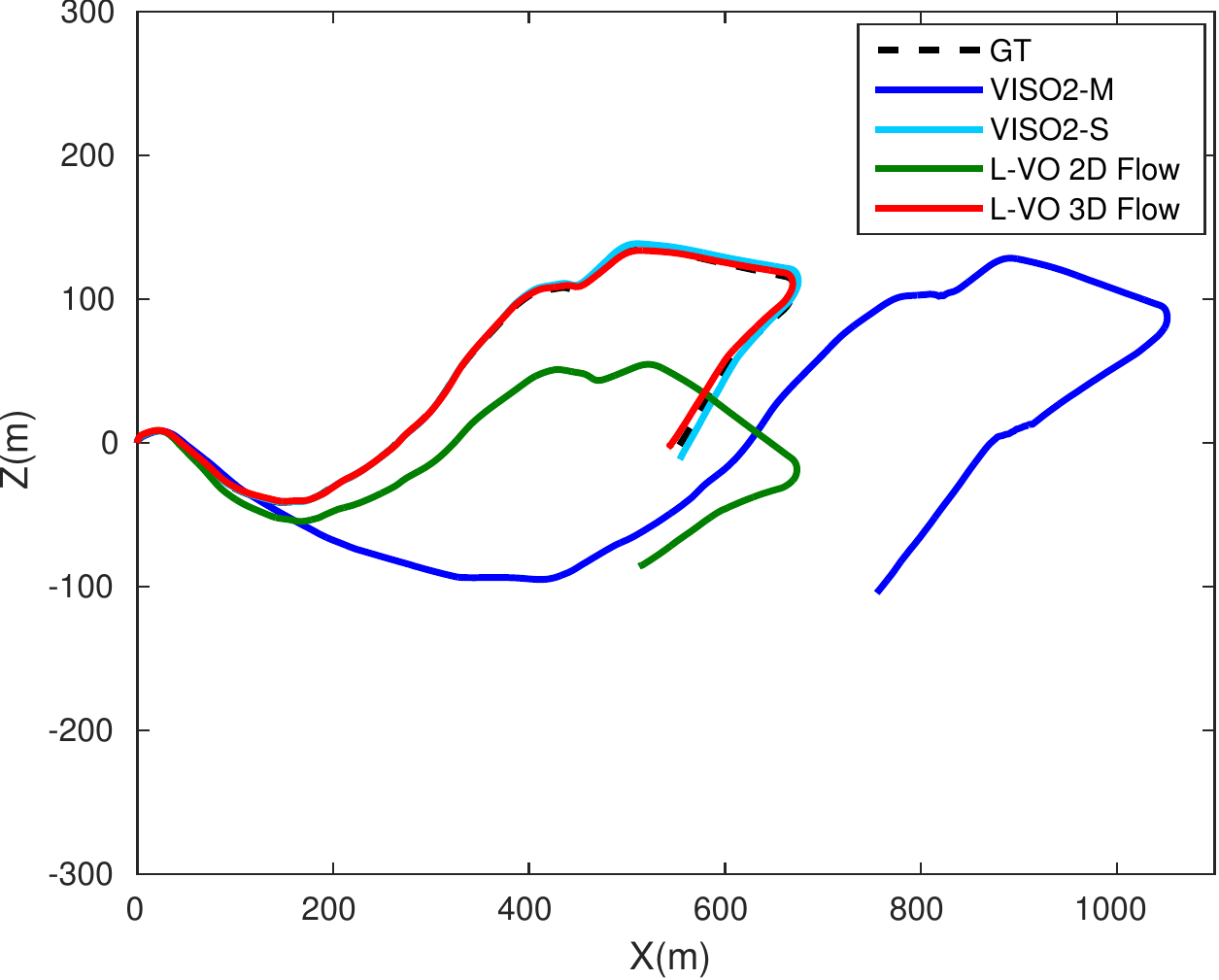}}
\caption{The predicted trajectories of the proposed L-VO Net on Sequences $03$, $04$, $05$, $06$, $07$ and $10$. The network is trained on Sequence $00$, $02$, $08$ and $09$.}
\label{fig:odometry}
\end{figure*}

\begin{table*}[thpb]
\small
\caption{The comparison of the performance of L-VO against the baselines on the KITTI dataset according to the evaluation method~\cite{wang2017end}. Note that VISO-S is a stereo VO and the other methods are monocular VO. The L-VO model is trained on the sequences $00$, $02$, $08$ and $09$, and evaluated on the rest.}
\centering
\resizebox{2\columnwidth}{!}{
\begin{tabular}{ c | c  c  c  c  c  c  c  c  c  c}
\toprule
\multirow{3}{*}{Seq.} & \multicolumn{2}{c}{VISO-S\cite{geiger2011stereoscan}} & \multicolumn{2}{c}{VISO-M\cite{geiger2011stereoscan}} & \multicolumn{2}{c}{ESP-VO\cite{wang2017end}} & \multicolumn{2}{c}{L-VO(2D Flow)} & \multicolumn{2}{c}{L-VO(3D Flow)}\\

& \multicolumn{2}{c}{$(1242\times 376)$} & \multicolumn{2}{c}{$(1242\times 376)$} & \multicolumn{2}{c}{$(1242\times 376)$} & \multicolumn{2}{c}{$(320\times 92)$} & \multicolumn{2}{c}{$(320\times 92)$}\\

& $t_{rel}(\%)$ & $r_{rel}(^\circ)$ & $t_{rel}(\%)$ & $r_{rel}(^\circ)$ & $t_{rel}(\%)$ & $r_{rel}(^\circ)$ & $t_{rel}(\%)$ & $r_{rel}(^\circ)$ & $t_{rel}(\%)$ & $r_{rel}(^\circ)$ \\
\midrule
03 & 1.71 & 1.12 & 9.02  & 2.83  & 6.72 & 6.46  & 3.35 & 1.62 & \textbf{3.18} & \textbf{1.31}\\
04 & 1.54 & 0.84 & 4.33  & 1.63  & 6.33 & 6.08  & 4.15 & 2.53 & \textbf{2.04} & \textbf{0.81}\\
05 & 2.36 & 1.20 & 19.16 & 3.62  & 3.35 & 4.93  & \textbf{2.49} & 1.19 & 2.59 & \textbf{0.99}\\
06 & 1.47 & 0.87 & 6.64  & 1.96  & 7.24 & 7.29  & 3.19 & 1.54 & \textbf{1.39} & \textbf{0.95}\\
07 & 2.37 & 1.78 & 26.54 & 5.92  & 3.52 & 5.02  & 17.2 & 10.4 & \textbf{2.81} & \textbf{2.54}\\
10 & 1.51 & 1.15 & 48.29 & 3.43  & 9.77 & 10.2  & 7.24 & \textbf{3.06} & \textbf{4.38} & 3.12\\
\midrule
Mean & 1.83 & 1.16 & 19.00 & 3.23 & 6.15 & 6.66 & 6.27 & 3.39 & \textbf{2.73} & \textbf{1.62}\\ 
\midrule 

\multicolumn{11}{c}{$t_{rel}$ and $r_{rel}$ are average translational RMSE(\%) and rotational RMSE($^\circ$/100m) over $100m-800m$ intervals.} \\ 

\bottomrule
\end{tabular}}

\label{table:The comparison of performance}
\end{table*}

\section{Experiments}\label{sec:4}

\begin{figure}[thpb]
\centering  
\subfigure[Translation vs path length. \label{fig:TP}]{\includegraphics[width= 0.23\textwidth]{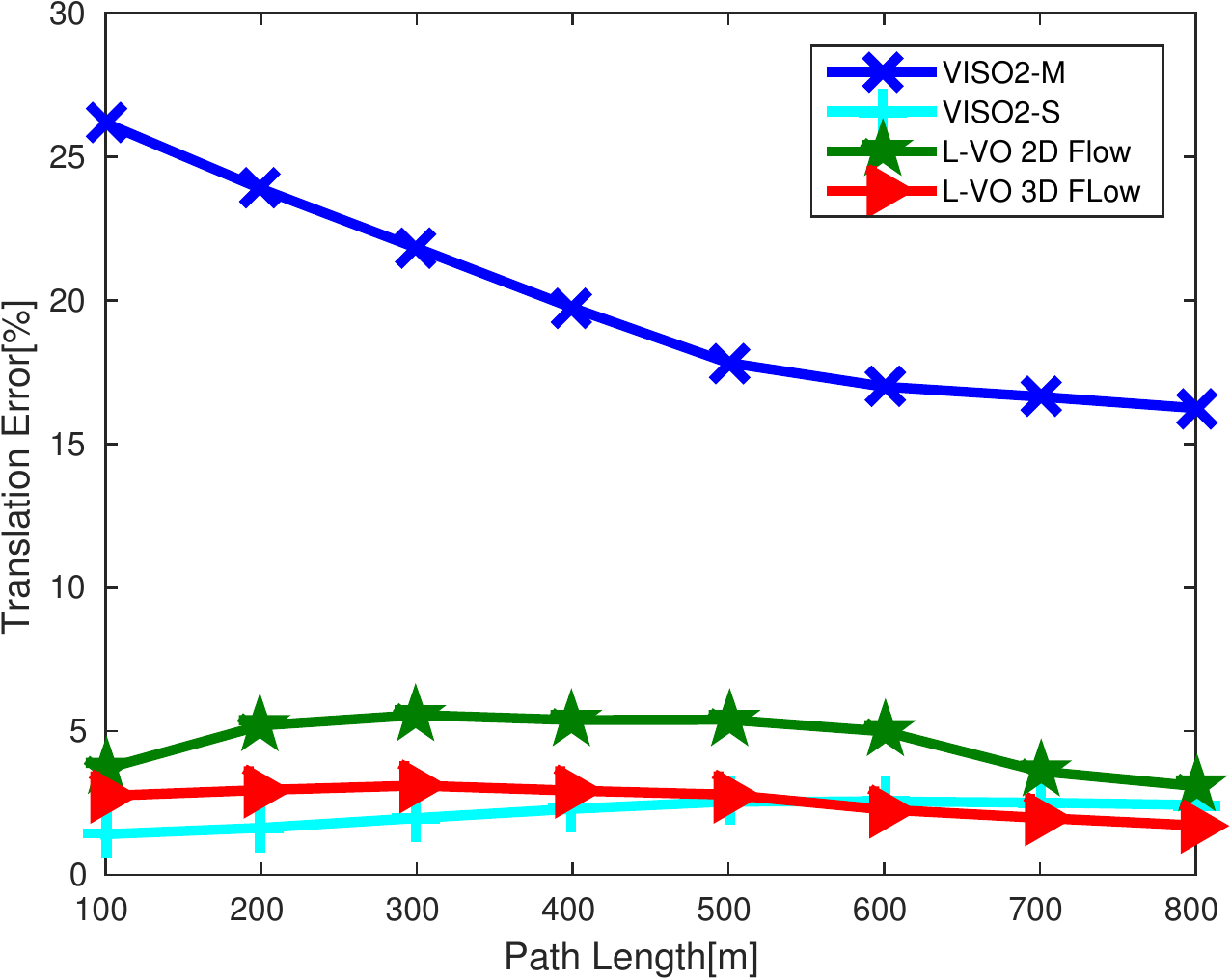}}
\subfigure[Rotation vs path length. \label{fig:RP}]{\includegraphics[width= 0.235\textwidth]{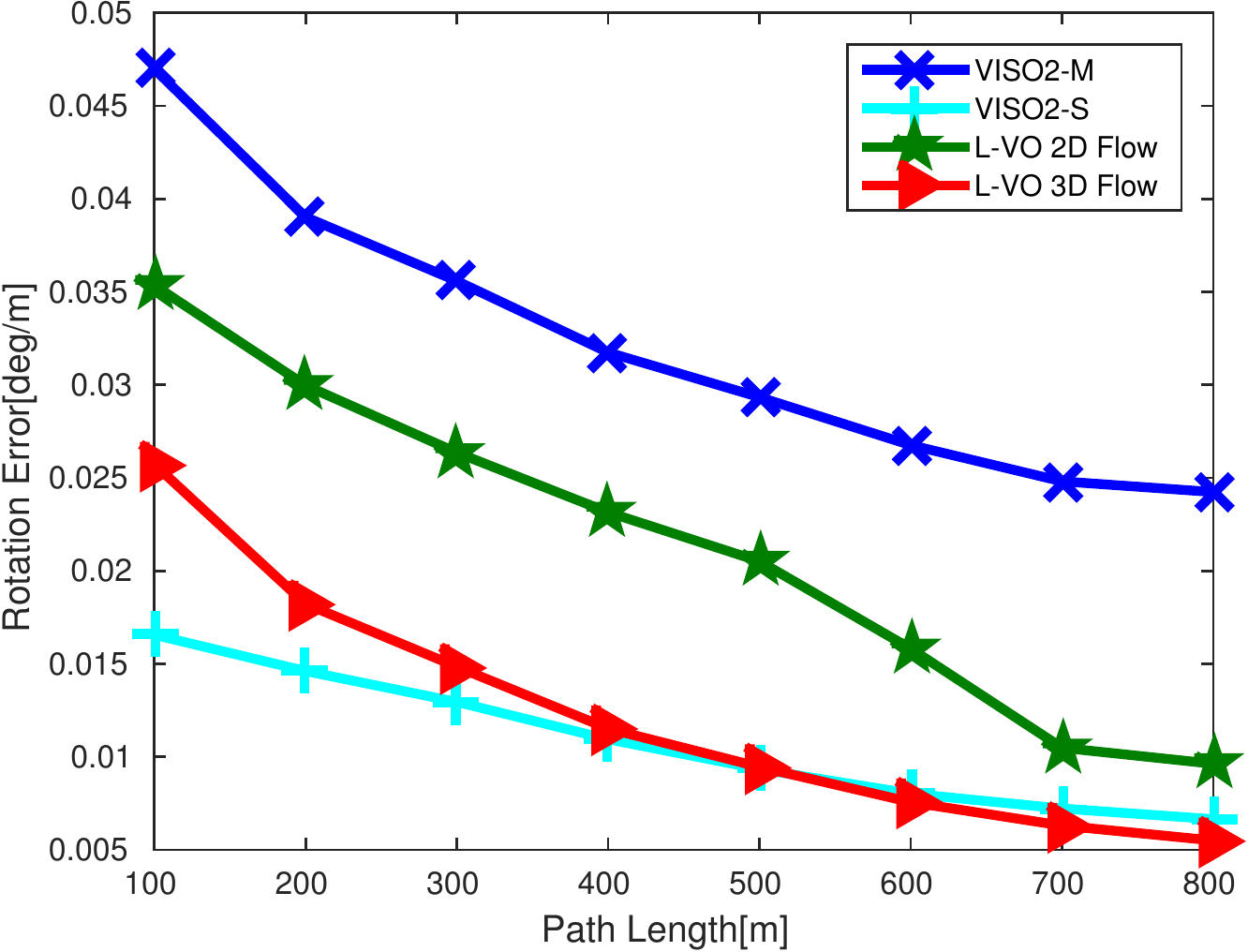}}
\subfigure[Translation vs speed. \label{fig:TS}]{\includegraphics[width= 0.23\textwidth]{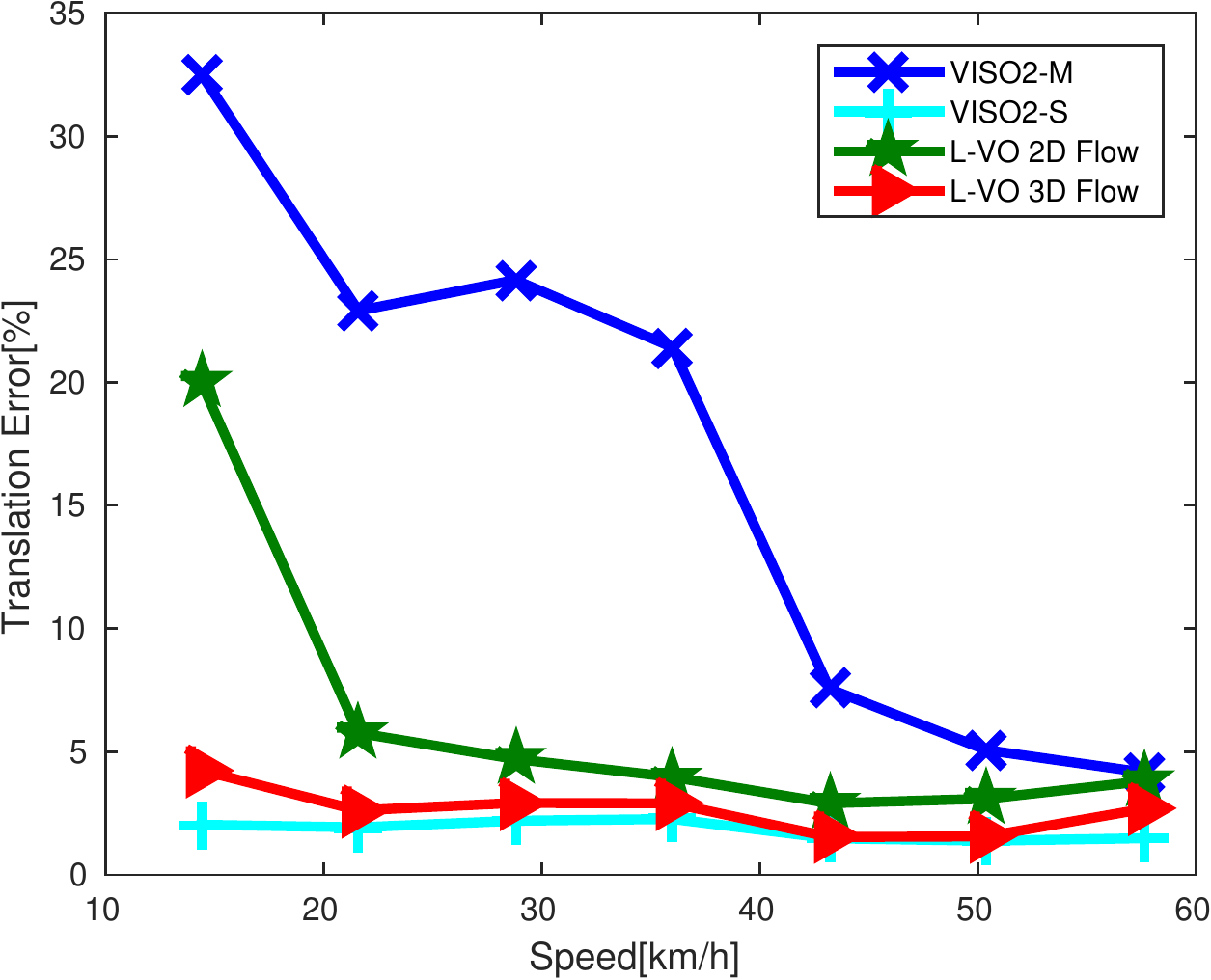}}
\subfigure[Rotation vs speed. \label{fig:RS}]{\includegraphics[width= 0.235\textwidth]{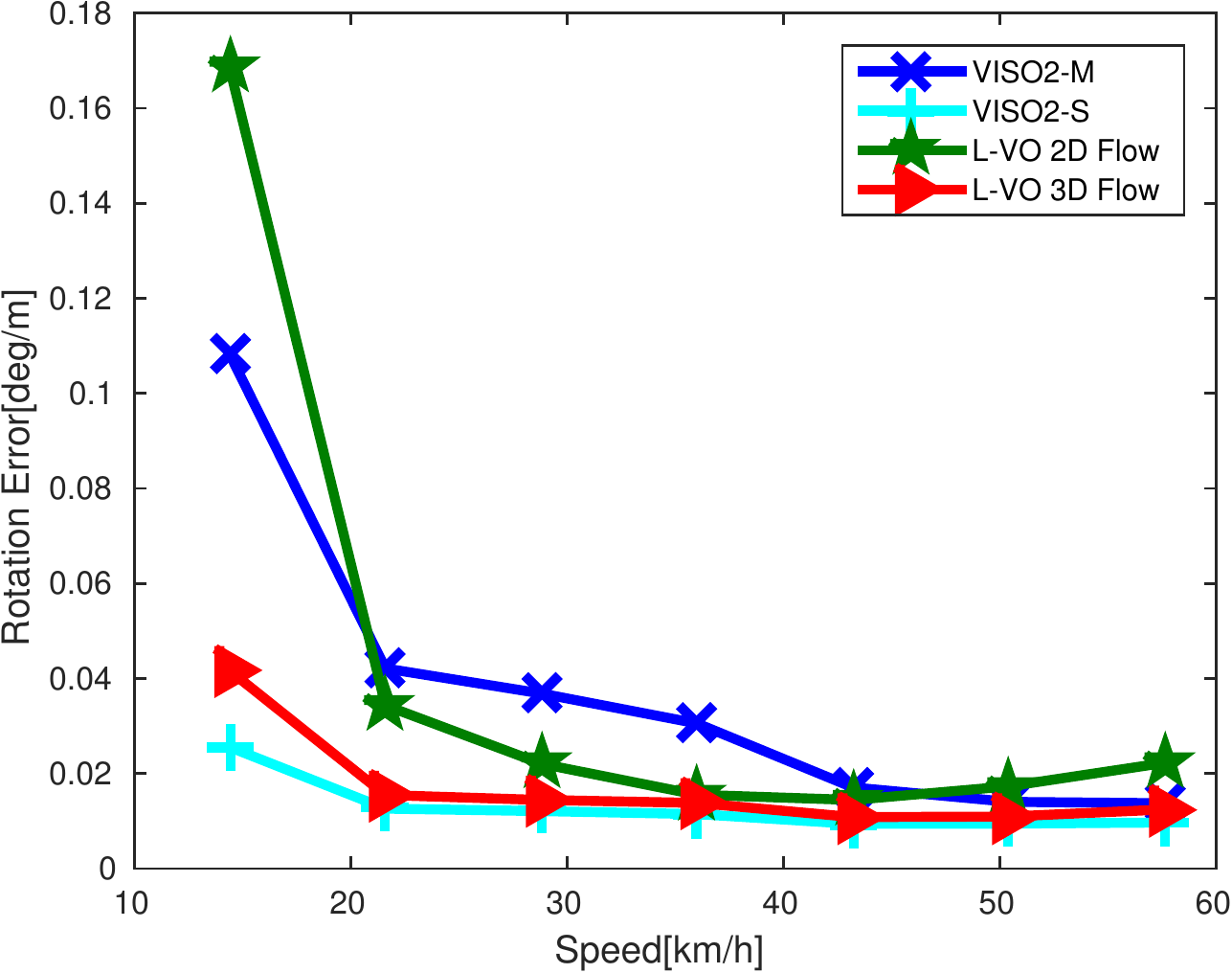}}
\caption{Average translational and rotational errors of the baselines against different path lengths and speeds. The L-VO model is trained on the sequences $00$, $02$, $08$ and $09$, and evaluated on the rest.}
\label{fig:plot error}
\end{figure}


\subsection{Dataset}\label{sec:4.1}
The proposed L-VO Net is evaluated on the most popular KITTI VO/SLAM benchmark. The KITTI VO/SLAM benchmark consists of $22$ sequences saved in PNG format. Sequences $00-10$ provide the sensor data with the accurate ground truth ($<10cm$) from a GPS/IMU system, while sequences $11-21$ only provide the raw sensor data. The large number of dynamic objects such as cars means that visual odometry could easily fail on this challenging dataset. 

\subsection{Network training and testing}\label{sec:4.2}
The network is trained with Adam optimization. The batch size is set to $100$, the momentum is fixed to $(0.9, 0.999)$, and the starting learning rate is $0.0001$. The step learning policy is adopted and the learning rate decay is set to $0.95$. The network is trained by $100$ epochs. In order to reduce the GPU memory requirement and training time, the raw images from the KITTI dataset are down-sampled $4$ times to $320 \times 96$. But using this small image size for training can definitely degrade the performance. The whole network is end-to-end trainable. Considering the GPU limitation, training the network step-by-step is more practicable. The pre-trained model~(without training on KITTI dataset) from~\cite{ilg2017flownet} and~\cite{godard2017unsupervised} is adapted and then fine-tuned using the training KITTI data~(as described in \ref{sec:4.3}). In order to enhance the performance and avoid over-fitting, both geometric augmentation (translation, rotation, scaling) and image augmentation (color, brightness, gamma) are employed. As mentioned in~\cite{dosovitskiy2015flownet}\cite{godard2017unsupervised} and \cite{wang2017end}, we also observe that these data augmentation techniques are crucial to improve the 2D flow estimation, depth prediction, and especially VO prediction, because of the limited number of training examples. During testing, the number of Gaussian samples is set to $10000$.   

\subsection{Visual odometry performance}\label{sec:4.3}
\begin{figure}[thpb]
	\centering
	\includegraphics[width= 0.3\textwidth]{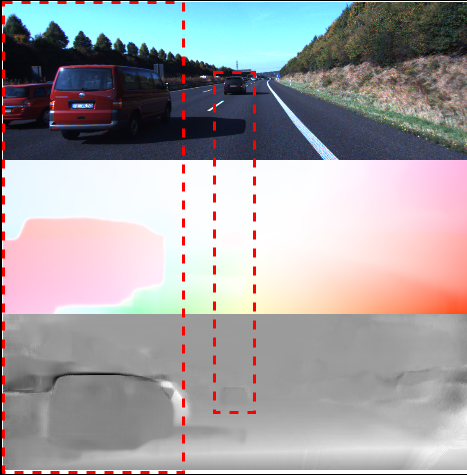}
	\caption{Sample image, 2D flow and depth flow (from top to bottom) from sequence 21 of KITTI. The red box refers to fast moving objects. This sequence is a very long distance scenario with many dynamic objects.}
	\label{fig:Failed}
\end{figure}

We perform two kinds of evaluation for the proposed methods. The first evaluation is based on sequence $00-10$. Both the qualitative and quantitative results are reported for analysis. For fair comparison, we follow the same partition proposed by~\cite{wang2017deepvo}\cite{wang2017end} and split the sequences 00-10 to 00, 02, 08, 09 for training and 03, 04, 05, 06, 07, 10 for testing. The second evaluation is based on sequence 00-21. The sequence 00-10 is employed for training and sequence 11-21 for testing. Only the qualitative results are provided because the ground truth of sequence 11-21 are not provided. The open-source visual odometry library VISO2~\cite{geiger2011stereoscan} is employed as the baseline method. It provides both monocular visual odometry and stereo odometry. For monocular VO, the fixed height (1.7) and pitch (-0.03) are employed in order to recover the absolute scale. 

We evaluate the learning odometry using the KITTI VO evaluation metrics, computing the average translational and rotational RMSE for all possible sub-sequences of length $(100,\ldots,800)$ meters. Note that the same evaluation metric was employed in~\cite{wang2017end}. 

For the first evaluation, the overall performance of average translational and rotational errors of L-VO based on 2D flow and 3D flow can reach $4.71\%, 0.0241 ^\circ/m$ and $2.68\%, 0.0143 ^\circ/m$, respectively, using the standard KITTI evaluation metrics. The detailed comparison of performances (some entries are copied from~\cite{wang2017end}) is shown in Table~\ref{table:The comparison of performance}. It is clear that the performance of both L-VO (2D) and L-VO (3D) is much better than conventional monocular VO. L-VO (3D) performs slightly worse than conventional stereo VO. This can also be seen in the predicted trajectory Fig.~\ref{fig:odometry}. Most of the time, the average drift distances of the red line (L-VO 3D) and green one (L-VO 2D) are between that of the light blue line (stereo VO) and dark blue line (monocular VO). The red line is much closer to the light blue line.

The main limitation of monocular VO and SLAM is the absolute scale estimation. However, with a deep learning method, the scale can be estimated more accurately without any scene-based geometric constraints such as camera height. This is one of the main reasons why the proposed L-VO(2D) and L-VO(3D) outperform the conventional monocular VO.

As we formulate VO prediction as a regression problem, multi-modal features can enhance the prediction. That is the reason why the result of L-VO(3D) is better than L-VO(2D) and closer to the performance of stereo VO. Another reason why L-VO(3D) can be close to stereo VO is that the $(x, z)$ constraint in the Bivariate Gaussian loss function can learn the translation correlation between the left/right and forward direction. This learned constraint can make the trajectory more accurate~-- see, for example, the straight line in Fig.~\ref{fig:sequence04} and corner in Fig.~\ref{fig:sequence05}. 

For low-speed scenarios, the magnitude of 2D flow is insignificant and thus provides a weak feature response to the network, while the magnitude of the depth flow is still quite strong even in a low-speed situation. Thus, the depth flow feature is a good complement to 2D flow in low-speed situations, which is further observed in Fig.~\ref{fig:TS} and Fig.~\ref{fig:RS}. Moreover, because the training data is only provided by $4$ sequences, multi-modal features, \emph{i.e.}, 3D flow can enhance the robustness of 6DOF relative pose regression. 

\begin{figure*}[thpb]
\centering
\subfigure[Sequence 11. \label{fig:sequence11}]{\includegraphics[width= 0.294\textwidth]{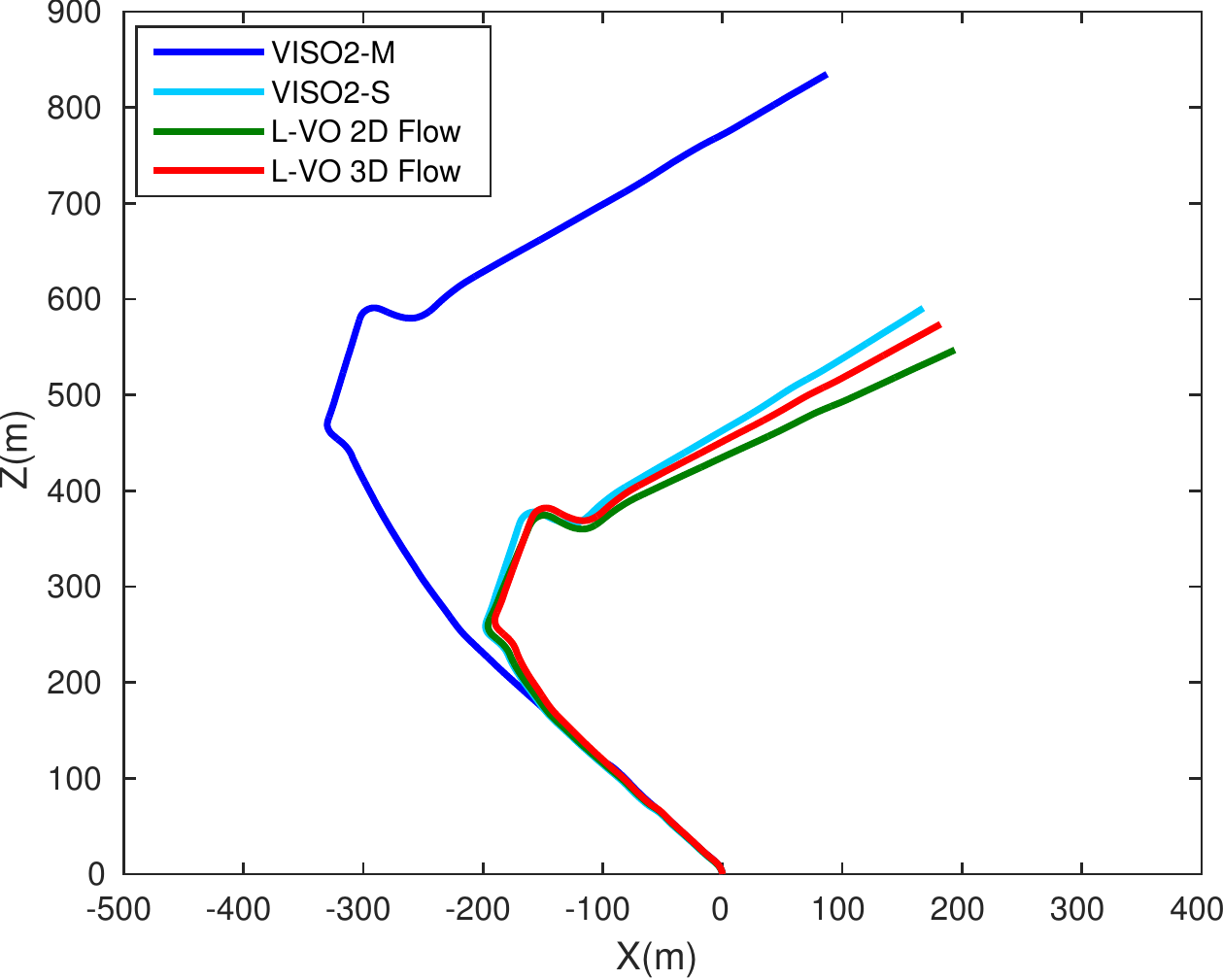}}
\subfigure[Sequence 12. \label{fig:sequence12}]{\includegraphics[width= \sizeOfimage\textwidth]{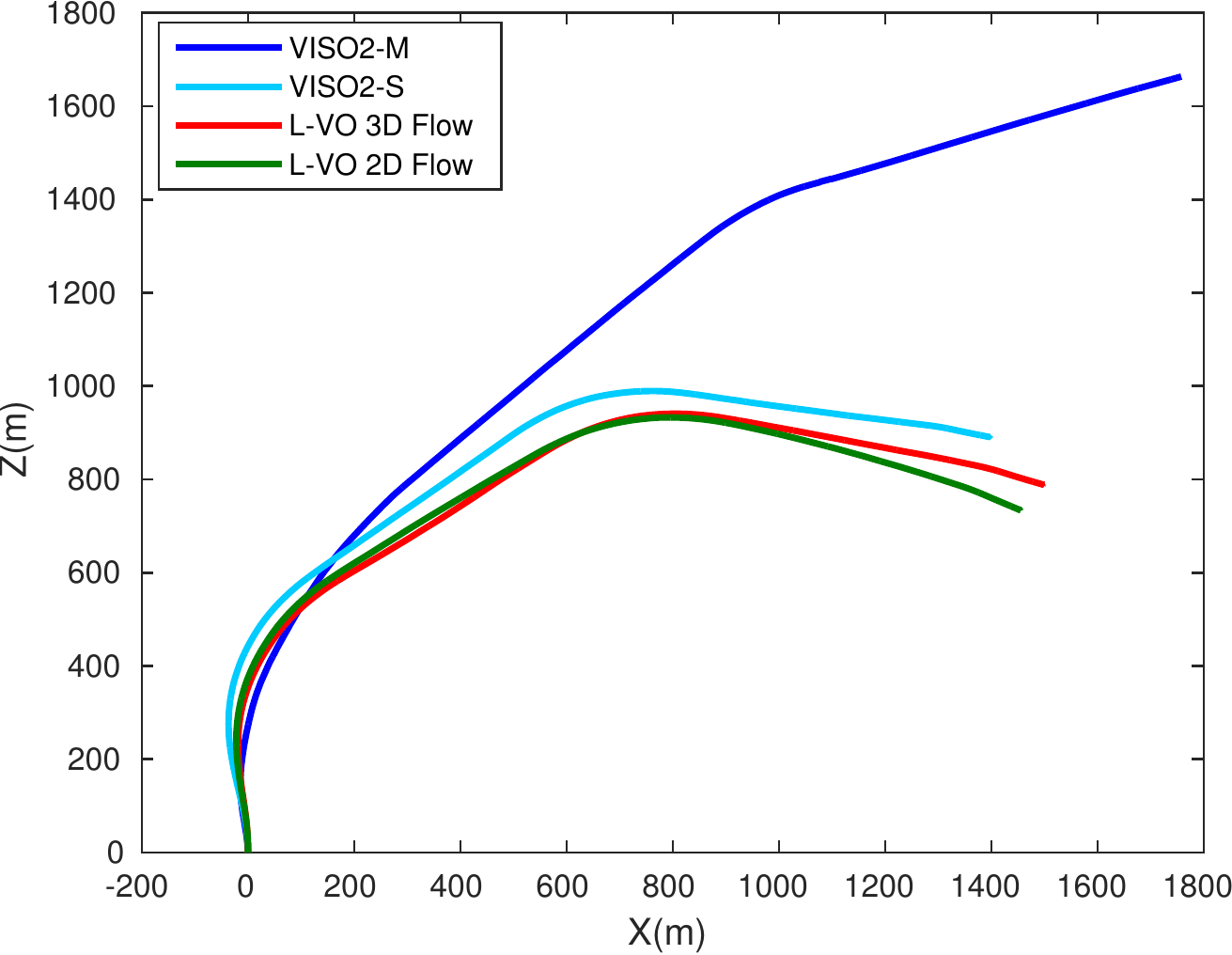}}
\subfigure[Sequence 14. \label{fig:sequence14}]{\includegraphics[width= \sizeOfimage\textwidth]{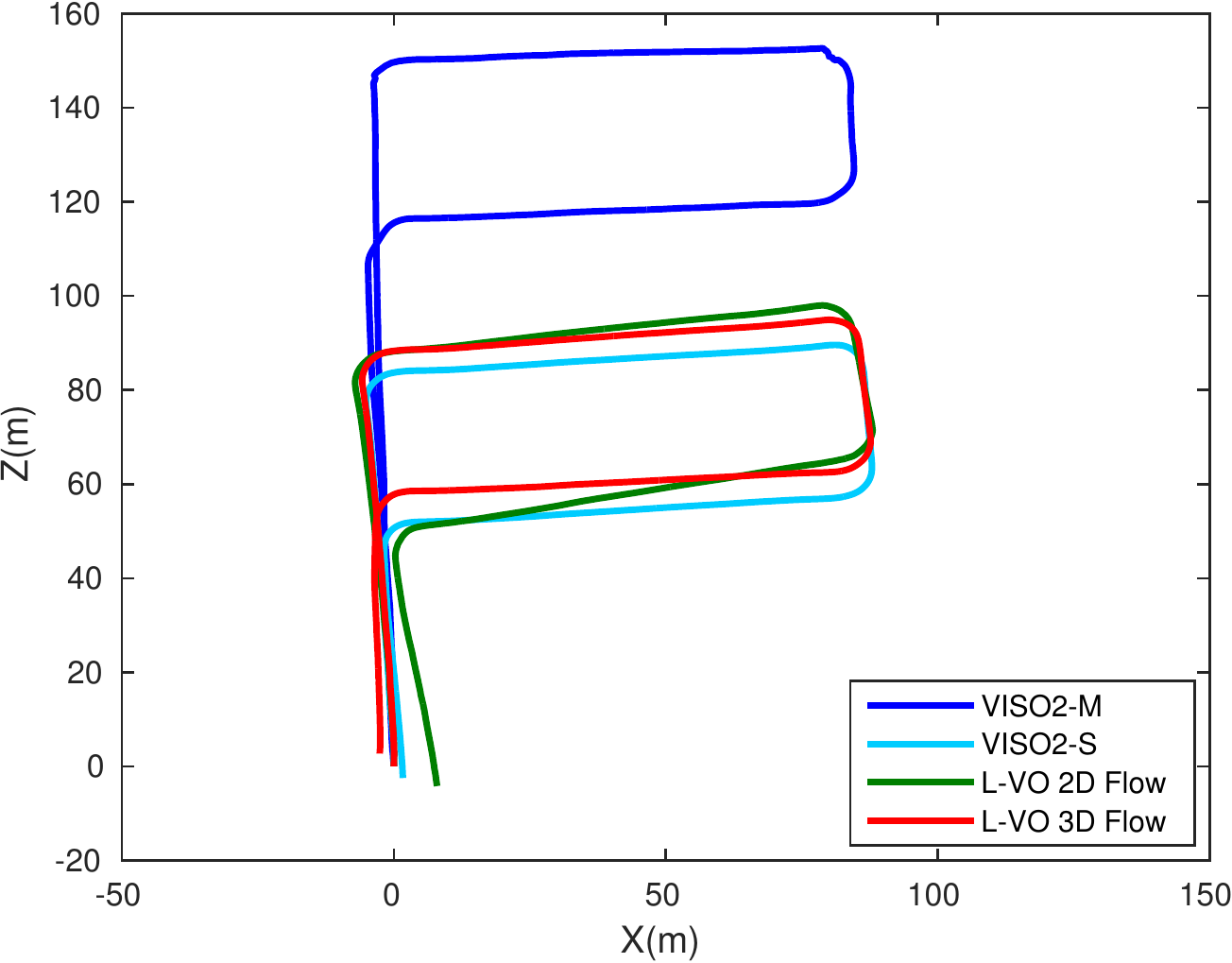}}
\subfigure[Sequence 15. \label{fig:sequence15}]{\includegraphics[width= \sizeOfimage\textwidth]{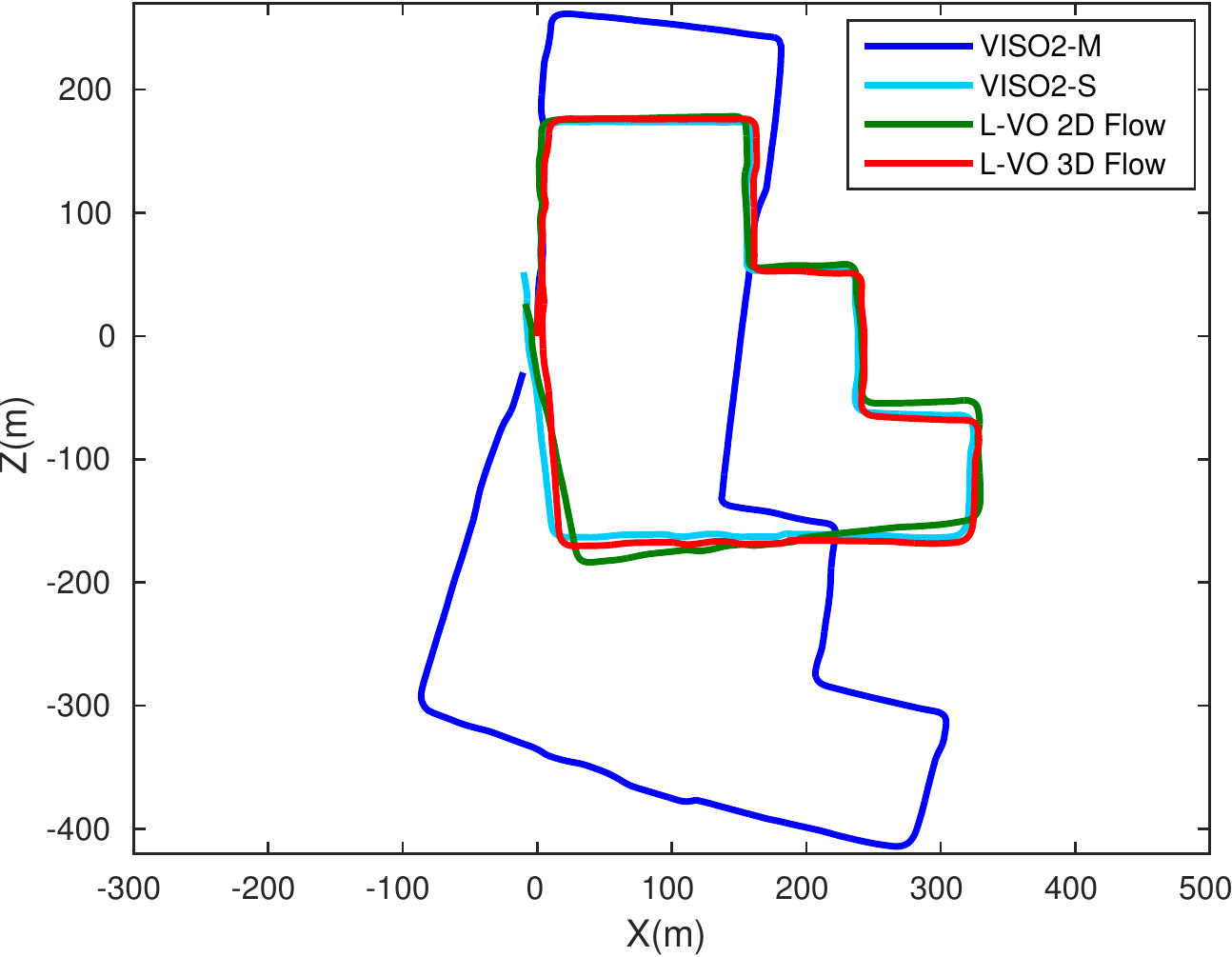}}
\subfigure[Sequence 17. \label{fig:sequence17}]{\includegraphics[width= \sizeOfimage\textwidth]{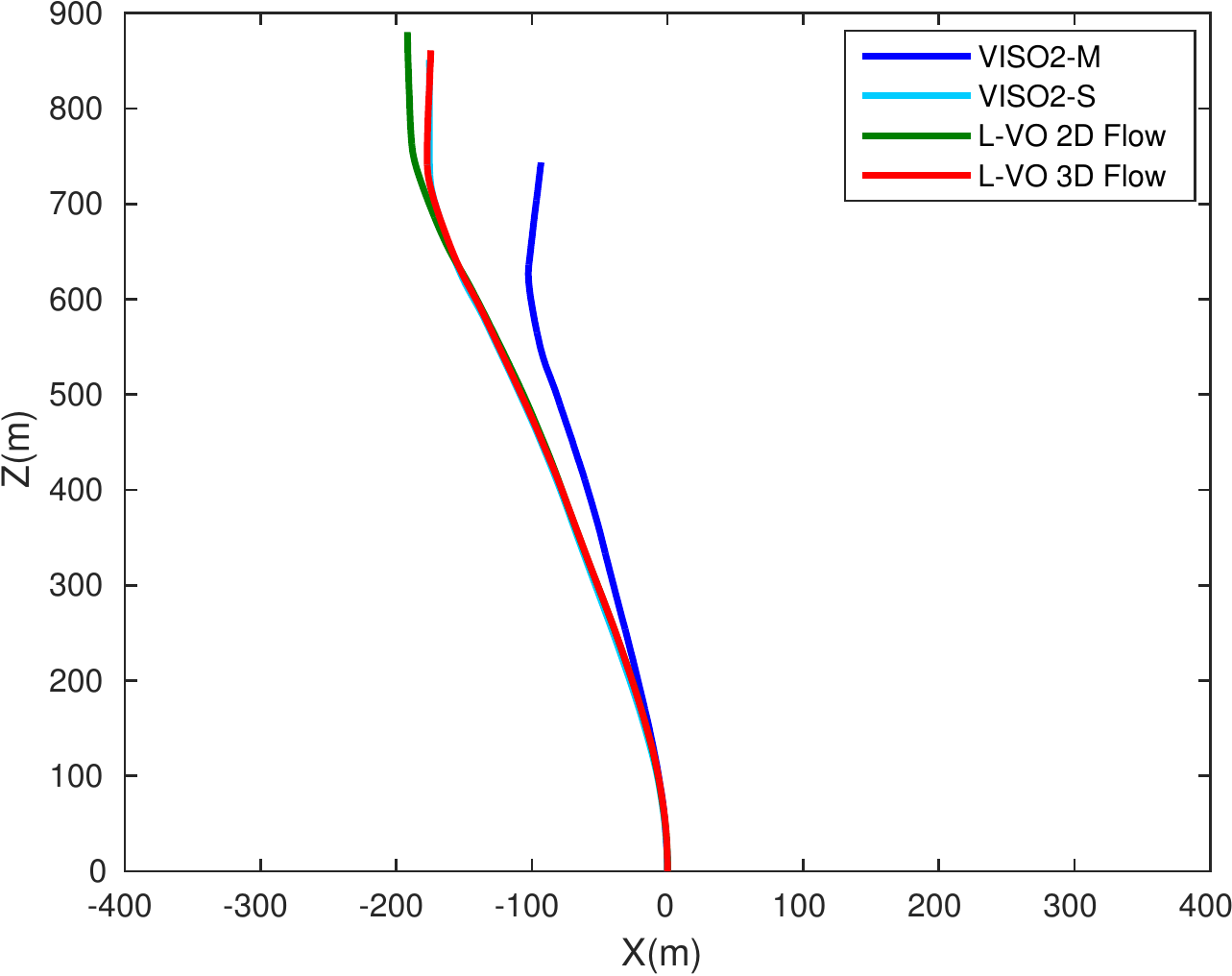}}
\subfigure[Sequence 18. \label{fig:sequence18}]{\includegraphics[width= \sizeOfimage\textwidth]{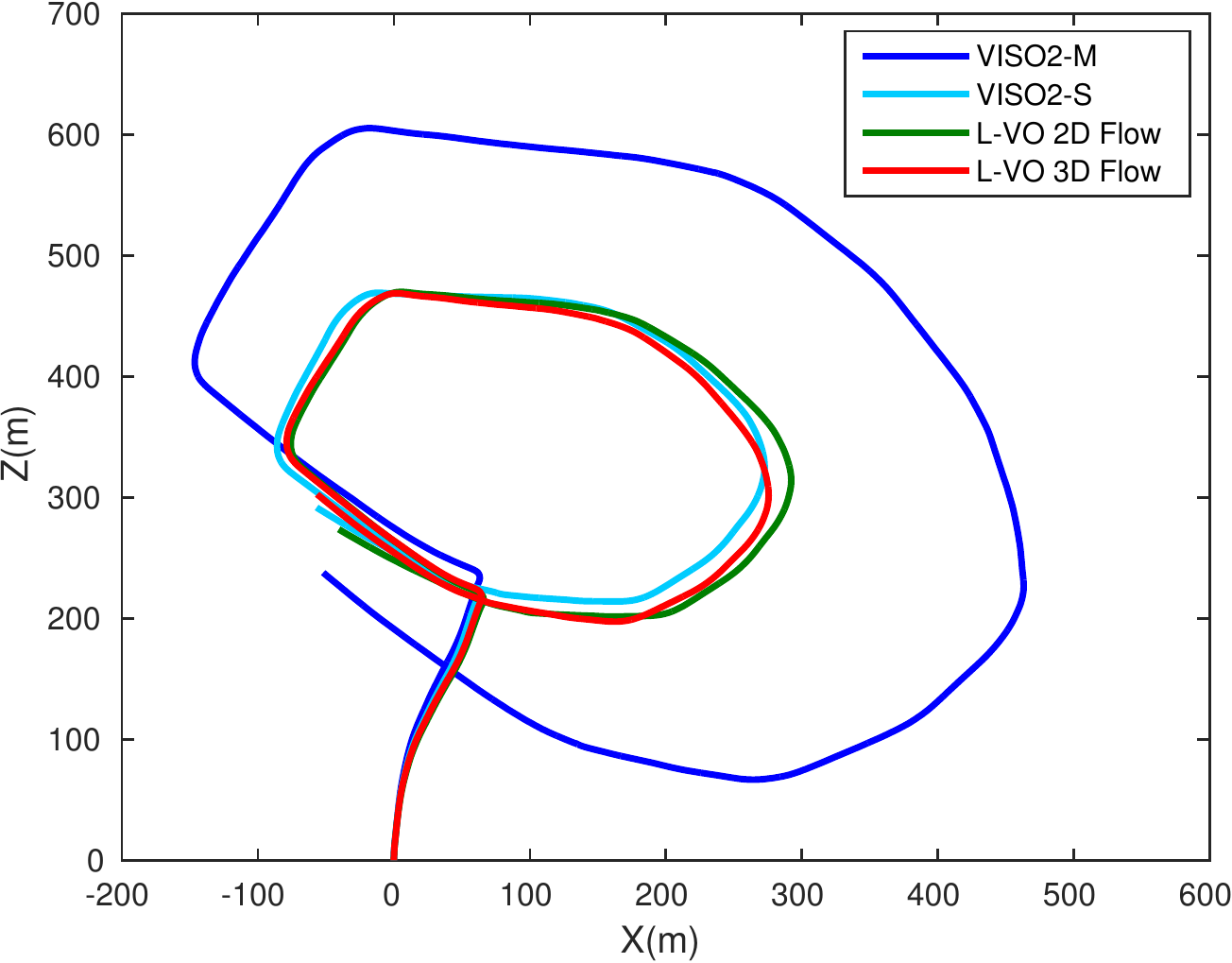}}
\subfigure[Sequence 19. \label{fig:sequence19}]{\includegraphics[width= \sizeOfimage\textwidth]{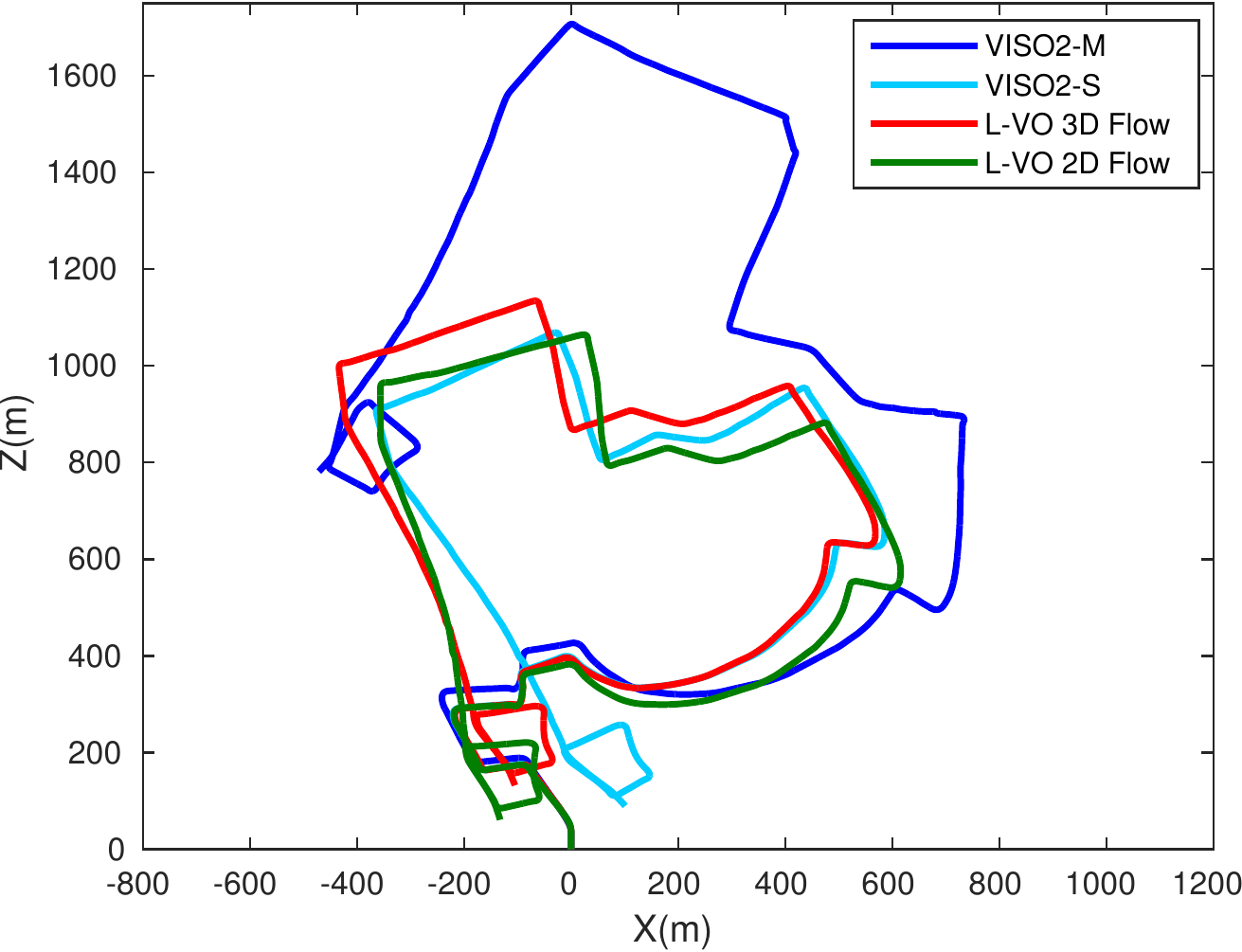}}
\subfigure[Sequence 20. \label{fig:sequence20}]{\includegraphics[width= 0.292\textwidth]{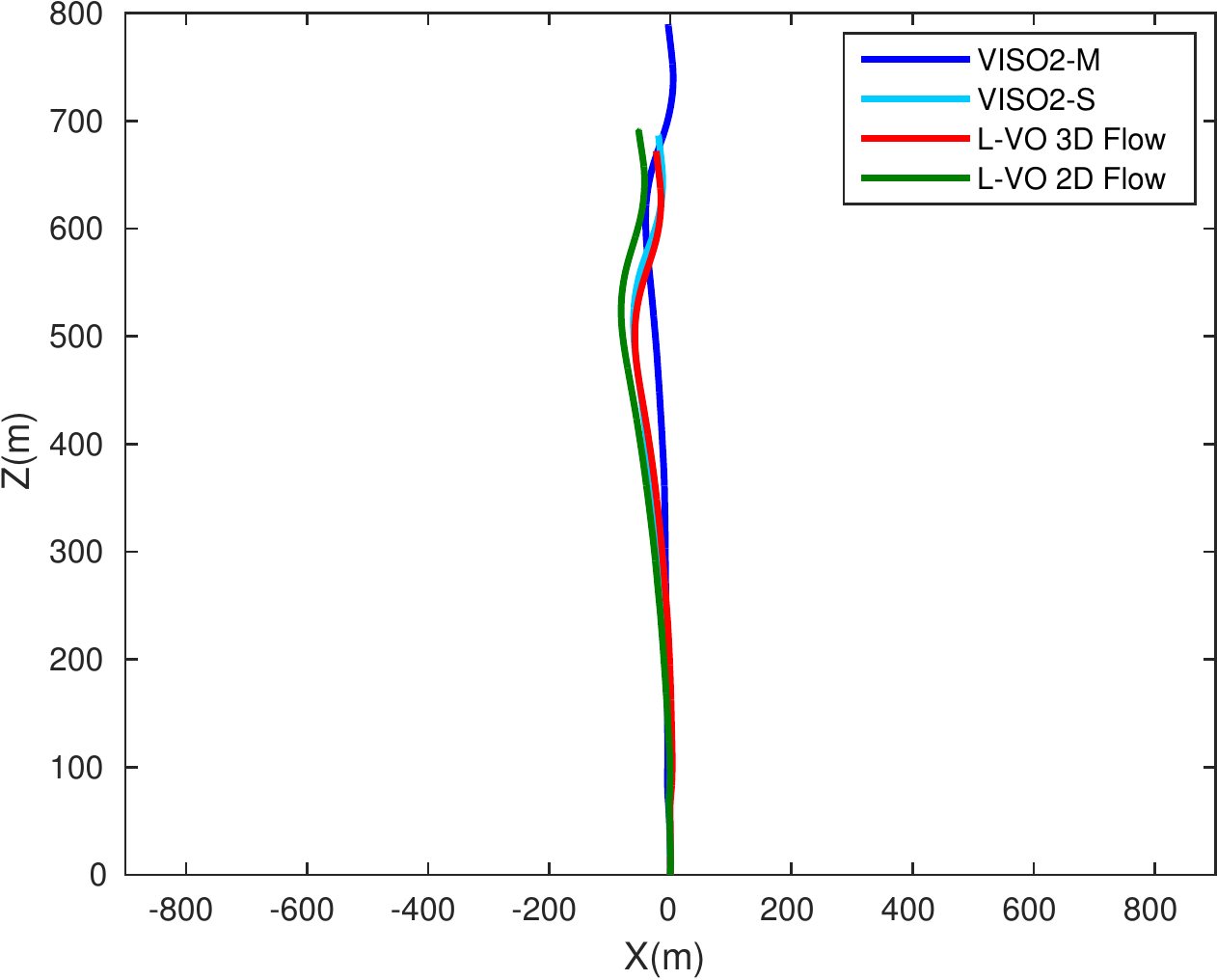}}
\subfigure[Sequence 21. \label{fig:sequence21}]{\includegraphics[width= \sizeOfimage\textwidth]{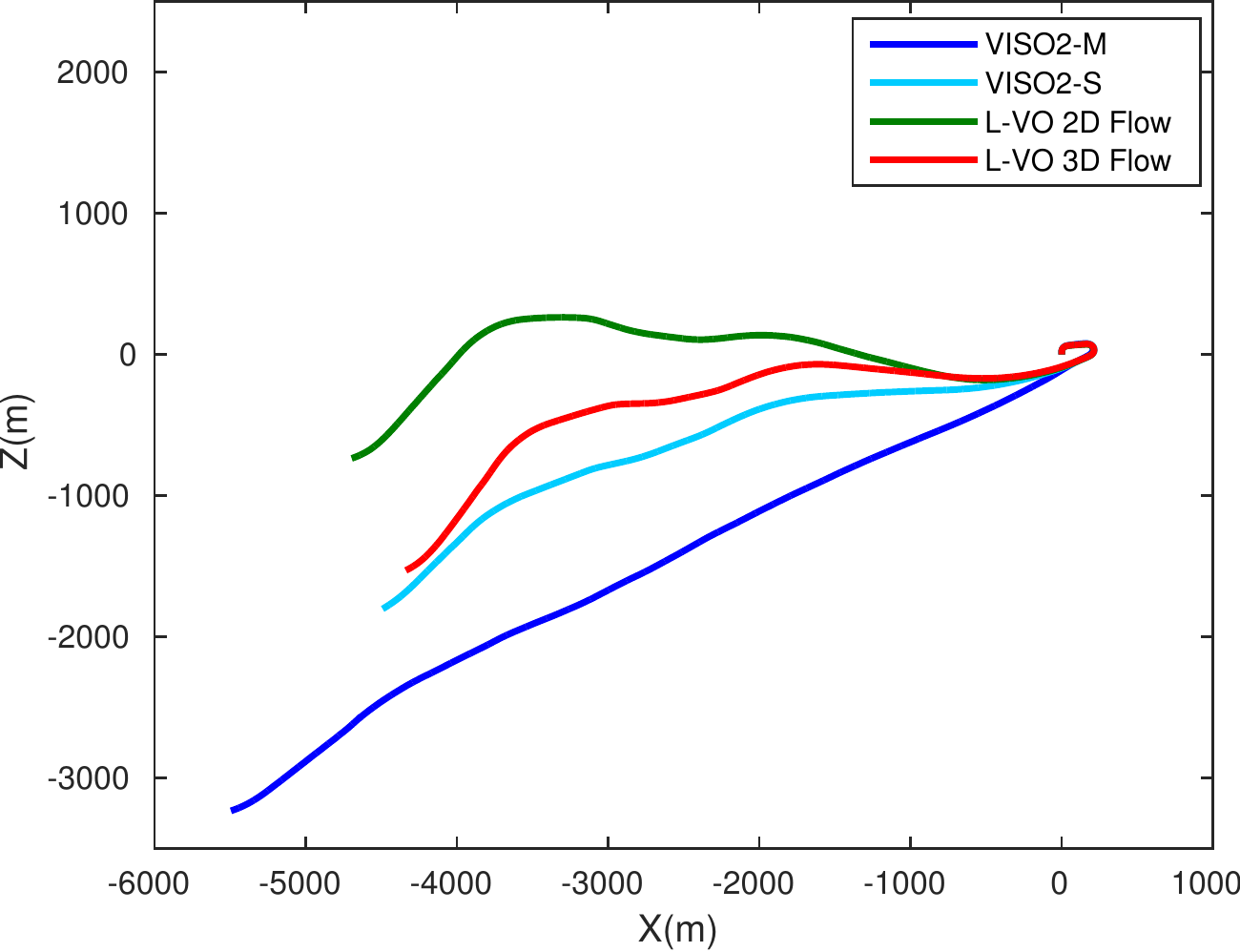}}
\caption{The predicted trajectories of proposed L-VO Net on Sequence 11, 12, 14, 15, 17, 18, 19, 20 and 21 (There are no ground truth available for these sequences). The L-VO model used is trained on Sequence from 00 to 10.}
\label{fig:odometry1}
\end{figure*}



For the second evaluation, the L-VO network is trained using more data, i.e.\ sequence 00-10. Due to the lack of ground truth, only qualitative results are shown in Fig.~\ref{fig:odometry1}. It can be seen that the L-VO network can also give a high-quality prediction in the new scenarios. Both L-VO(2D) and L-VO(3D) outperform monocular VO thanks to better scale estimation. The trajectory of L-VO(3D) is closer to stereo VO than L-VO(2D). However, the performance of L-VO(2D) is boosted more than L-VO(3D) by using more training data. 

During testing, we observe that L-VO cannot give a similar prediction to stereo VO for sequence 21 (Fig.~\ref{fig:sequence21}). This sequence is very challenging as it is captured over a long distance in a high-speed scenario (up to 80km/h). The main difficulty L-VO encounters is the high number of moving objects such as fast-moving cars in this street. As displayed in Fig.~\ref{fig:Failed}, the main flow feature is extracted from the fast-moving cars. Therefore, the main challenge for flow-based learning VO is to remove the effects of dynamic objects.  

\begin{figure*}[thpb]
	\centering
	\includegraphics[width= 0.8\textwidth]{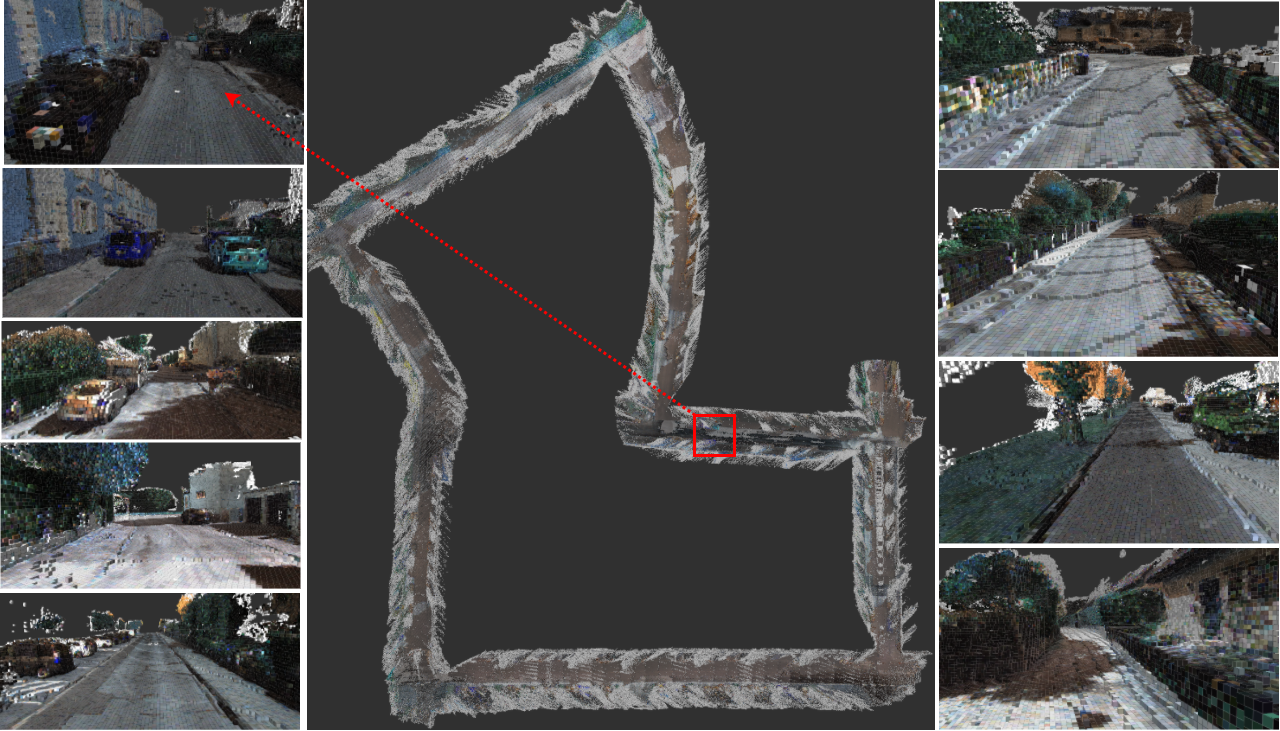}
	\caption{The center image is the global dense 3D map of sequence 07 in the KITTI dataset. The small images in the surrounding show enlarged local areas of the global map.}
	\label{fig:Map}
\end{figure*}

\subsection{Dense 3D mapping}\label{sec:4.4}
A learning monocular SLAM system integrated with L-VO(3D) is deployed in this paper. The whole system is implemented under ROS
and the neural network is implemented using Tensorflow
trained on an NVIDIA Titan GPU. Compared to LSD and ORB monocular SLAM, our system can generate a significantly denser 3D map. In order to alleviate the border blur and wrong prediction of depth, depth fusion from multiple frames is employed during mapping. In order to reduce the hardware resource requirement, OctoMap is used for the map representation instead of the point cloud. Given the dense refinement of depth information, a dense 3D map can be generated online. In Fig.~\ref{fig:Map}, the center image is the dense 3D map of the sequence $07$ in the KITTI dataset and the small images in the surrounding show enlarged local areas of the global map. It can be seen that after depth fusion, sharply defined shapes such as the car, trees and building are obtained. Moreover, a lot of outliers and noise are removed to make the map cleaner. 
\section{Conclusion}\label{sec:5}
In this paper, a learning system for monocular SLAM is proposed, which can deploy simultaneous localization using a L-VO neural network and the dense 3D mapping. Its performance exceeds most of the monocular SLAM approaches and is even comparable with some stereo SLAM approaches. Compared with conventional SLAM, its main limitations are the high computational requirements and 
high dataset bias. A demo can be found on the first author's Youtube channel\footnote{\url{https://youtu.be/Ccj1O7yndIk}}.  


\section{Acknowledgement}
The authors was funded by a DISTINCTIVE scholarship, EU H2020 projects: RoMaNS (645582) \& ILIAD (732737). 
\bibliographystyle{IEEEtran}
\bibliography{./references}
\end{document}